\documentclass[letterpaper]{article} 
\usepackage{aaai25}  
\usepackage{times}  
\usepackage{helvet}  
\usepackage{courier}  
\usepackage[hyphens]{url}  
\usepackage{graphicx} 
\usepackage{natbib}  
\usepackage{caption} 
\usepackage{algorithm}
\usepackage{algorithmic}

\usepackage{newfloat}
\usepackage{listings}
\DeclareCaptionStyle{ruled}{labelfont=normalfont,labelsep=colon,strut=off} 
\floatstyle{ruled}
\newfloat{listing}{tb}{lst}{}
\floatname{listing}{Listing}
\usepackage{subfig}
\usepackage{amsmath}
\usepackage{booktabs}
\newcommand{\vdashlineup}{\rotatebox{90}{
\rule{6pt}{1pt}\hspace{0.5em}\rule{6pt}{1pt}\hspace{0.5em}
\rule{6pt}{1pt}\hspace{0.5em}\rule{6pt}{1pt}\hspace{0.5em}
\rule{6pt}{1pt}\hspace{0.5em}\rule{6pt}{1pt}\hspace{0.4em}\rule{3pt}{1pt}}}
\newcommand{\vdashlinedown}{\rotatebox{90}{
\rule{6pt}{1pt}\hspace{0.5em}\rule{6pt}{1pt}\hspace{0.5em}
\rule{6pt}{1pt}\hspace{0.5em}\rule{6pt}{1pt}\hspace{0.5em}
\rule{6pt}{1pt}\hspace{0.5em}\rule{6pt}{1pt}\hspace{0.5em}
\rule{6pt}{1pt}\hspace{0.5em}\rule{6pt}{1pt}}}

\title{Move and Act: Enhanced Object Manipulation and Background Integrity for Image Editing}
\author{
    Pengfei Jiang\textsuperscript{\rm 1},
    Mingbao Lin\textsuperscript{\rm 2},
    Fei Chao\textsuperscript{\rm 1}\thanks{Corresponding author}
}
\affiliations{
    
    \textsuperscript{\rm 1}Key Laboratory of Multimedia Trusted Perception and Efficient Computing, \\
    Ministry of Education of China, Xiamen University, China \\
    \textsuperscript{\rm 2}Skywork AI Singapore\\
    
    pengfeijiang@stu.xmu.edu.cn, linmb001@outlook.com, fchao@xmu.edu.cn
}

\usepackage{bibentry}

\begin{document}

\maketitle

\begin{abstract}
Current methods commonly utilize three-branch structures of inversion, reconstruction, and editing, to tackle consistent image editing task. However, these methods lack control over the generation position of the edited object and have issues with background preservation. To overcome these limitations, we propose a tuning-free method with only two branches: inversion and editing. This approach allows users to simultaneously edit the object's action and control the generation position of the edited object. Additionally, it achieves improved background preservation. Specifically, we transfer the edited object information to the target area and repair or preserve the background of other areas during the inversion process at a specific time step. In the editing stage, we use the image features in self-attention to query the key and value of the corresponding time step in the inversion to achieve consistent image editing. Impressive image editing results and quantitative evaluation demonstrate the effectiveness of our method. The code is available at \url{https://github.com/mobiushy/move-act}.
\end{abstract}

%

\begin{figure}
  \centering
  \captionsetup[subfloat]{labelsep=none,format=plain,labelformat=empty}
  \subfloat[(a) Input]   
  {
      \label{fig:subfig1-1}\includegraphics[width=0.32\linewidth]{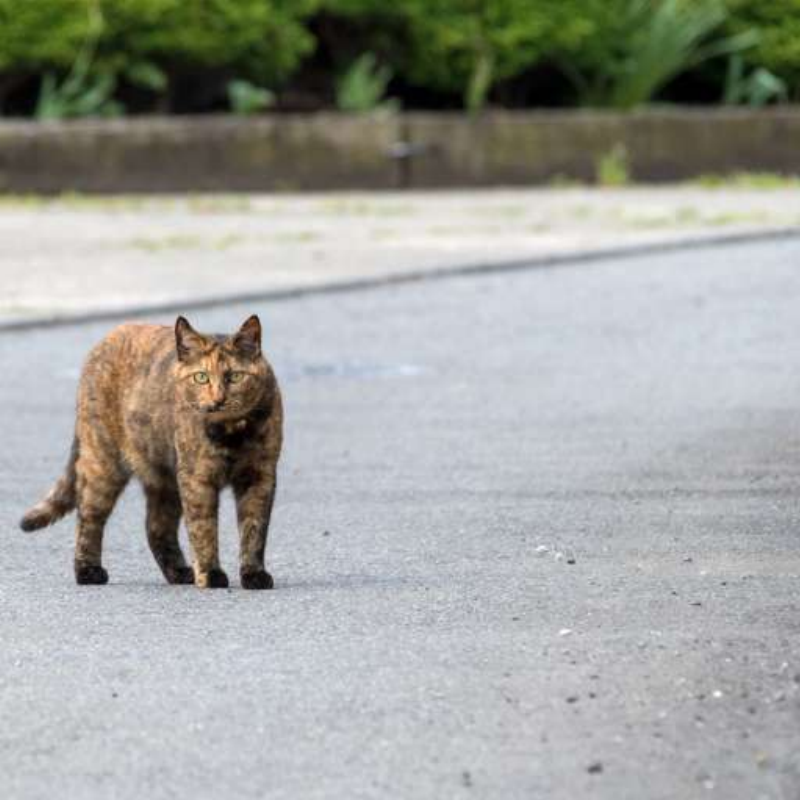}
  }
  \subfloat[(b) \textit{A running cat}]   
  {
      \label{fig:subfig1-2}\includegraphics[width=0.32\linewidth]{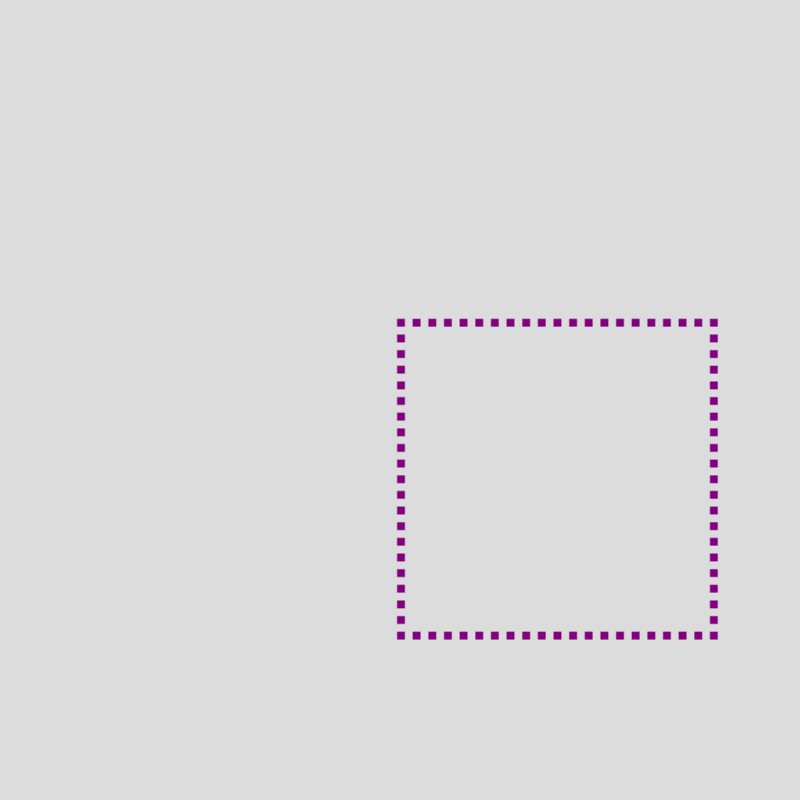}
      \label{fig:subfig1-3}\includegraphics[width=0.32\linewidth]{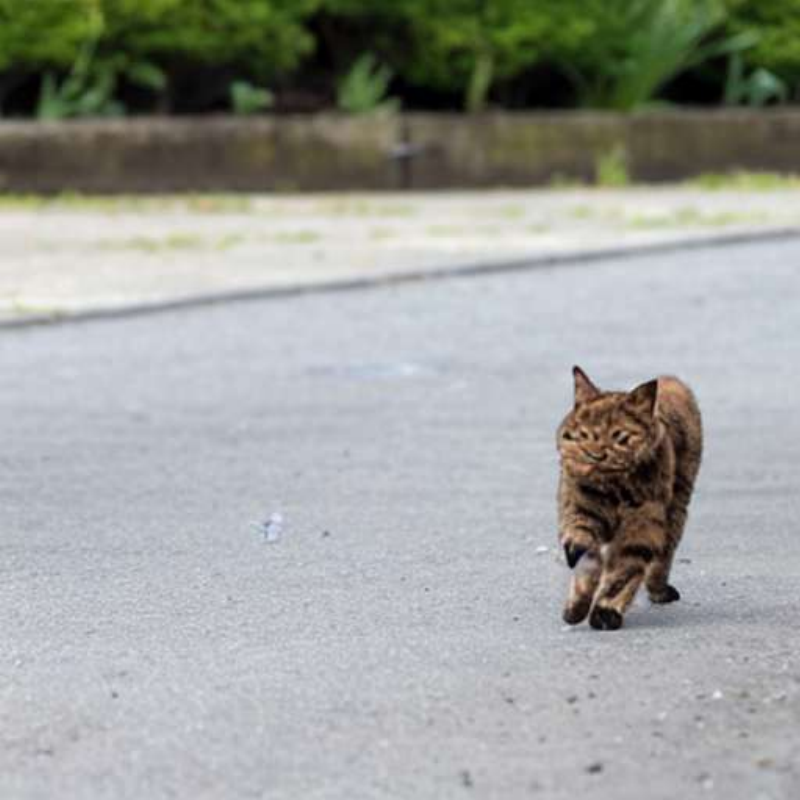}
  }
  \vspace{-5pt}
  \subfloat[(c) Input]   
  {
      \label{fig:subfig1-4}\includegraphics[width=0.32\linewidth]{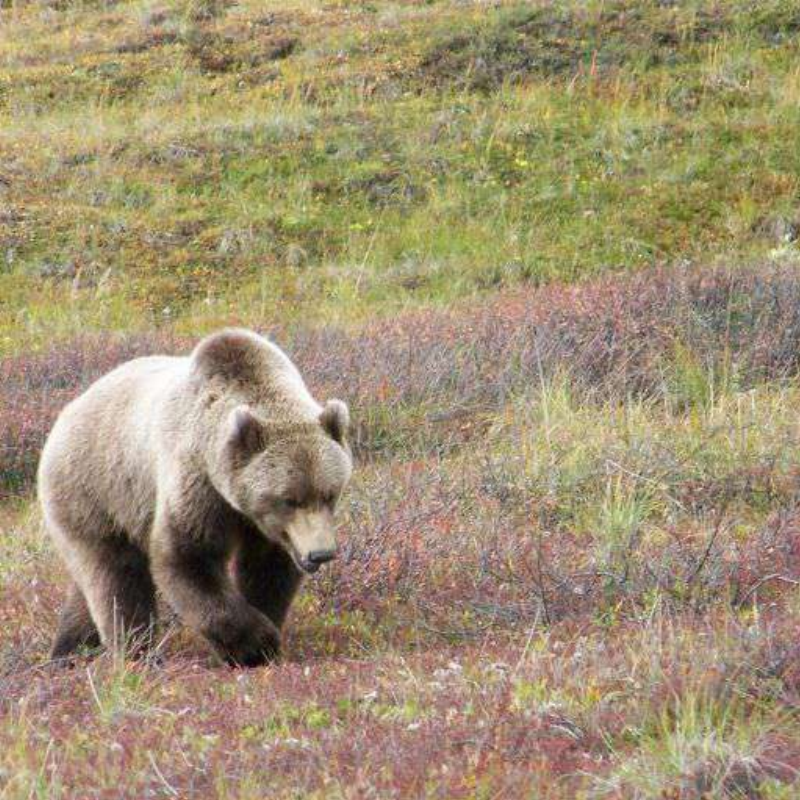}
  }
  \subfloat[(d) \textit{A running bear}]   
  {
      \label{fig:subfig1-5}\includegraphics[width=0.32\linewidth]{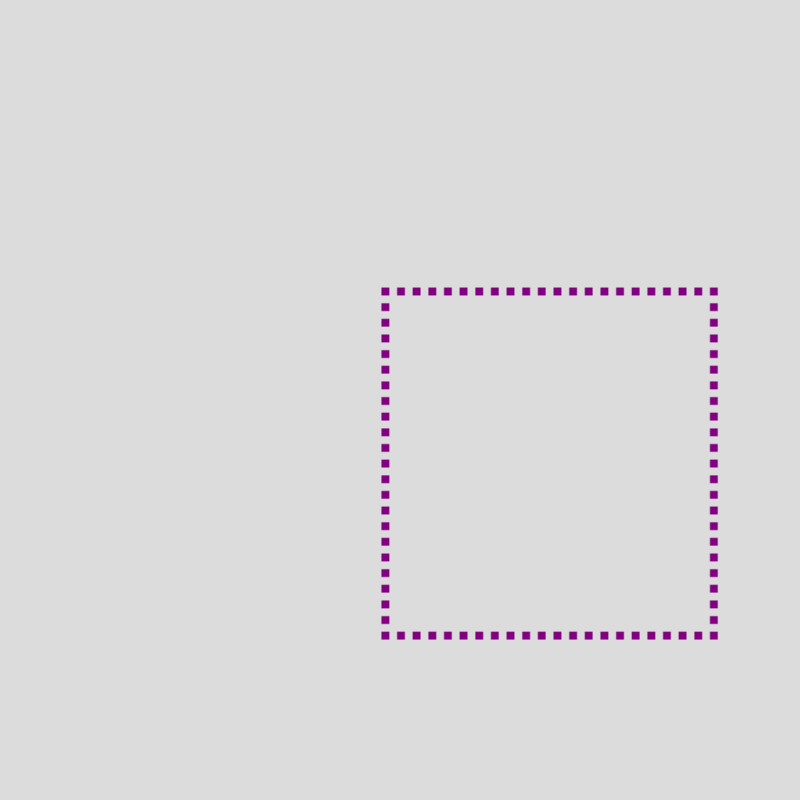}
      \label{fig:subfig1-6}\includegraphics[width=0.32\linewidth]{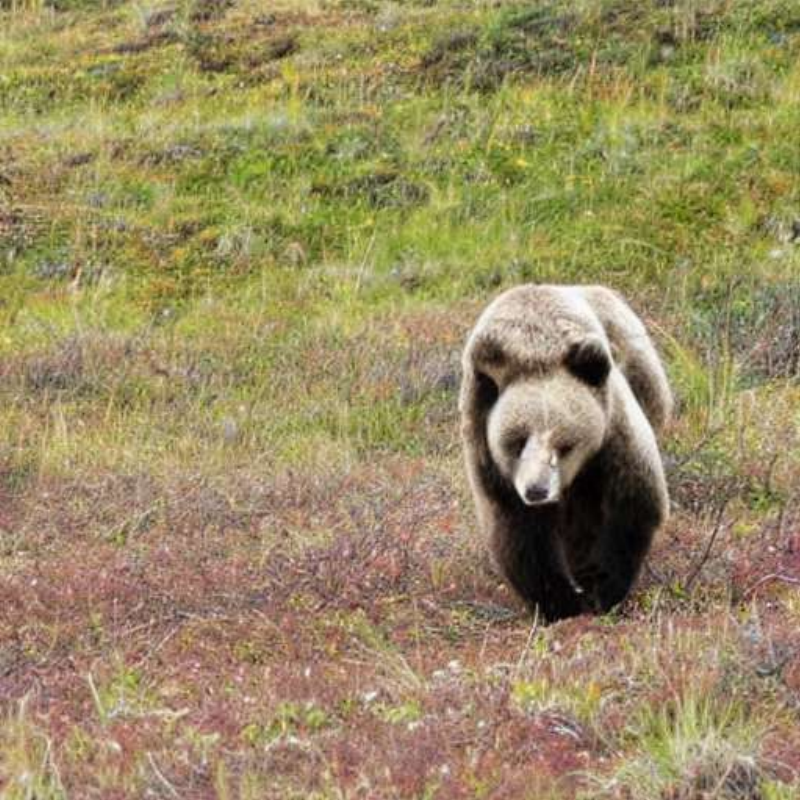}
  }
  \vspace{5pt}
  ~\noindent\rule{8.5cm}{0.7pt}
   \subfloat[\footnotesize\textit{Original image}]
  {
      \includegraphics[width=0.32\linewidth]{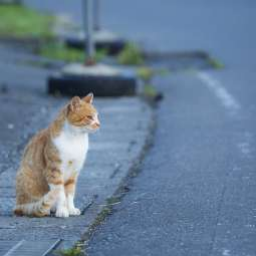}
  }
  \vdashlineup
  \subfloat[(f) InstructPix2Pix]
  {
      \includegraphics[width=0.32\linewidth]{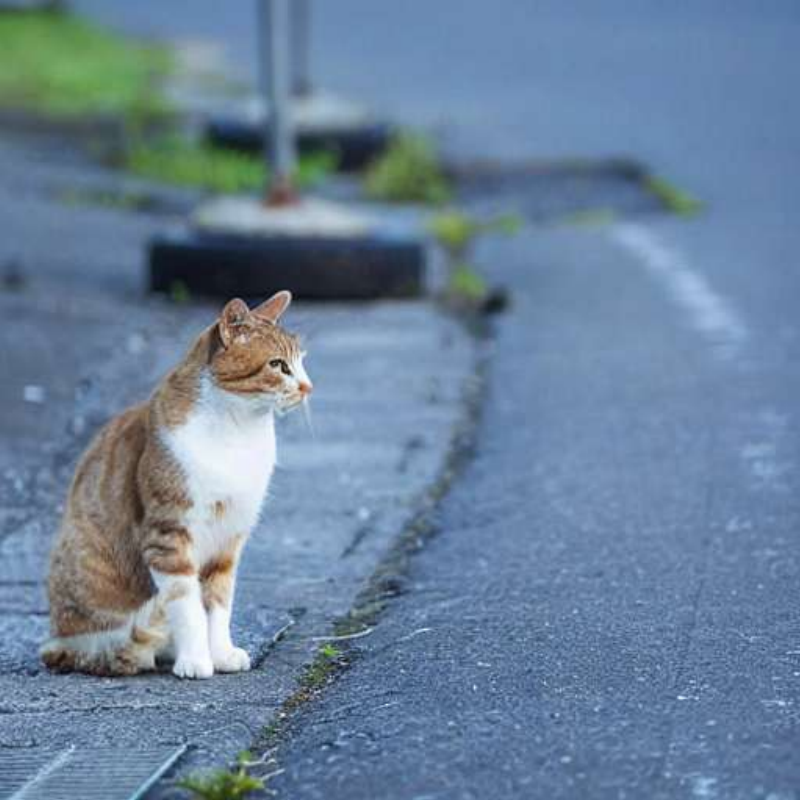}
  }
  \subfloat[(g) PnP]
  {
      \includegraphics[width=0.32\linewidth]{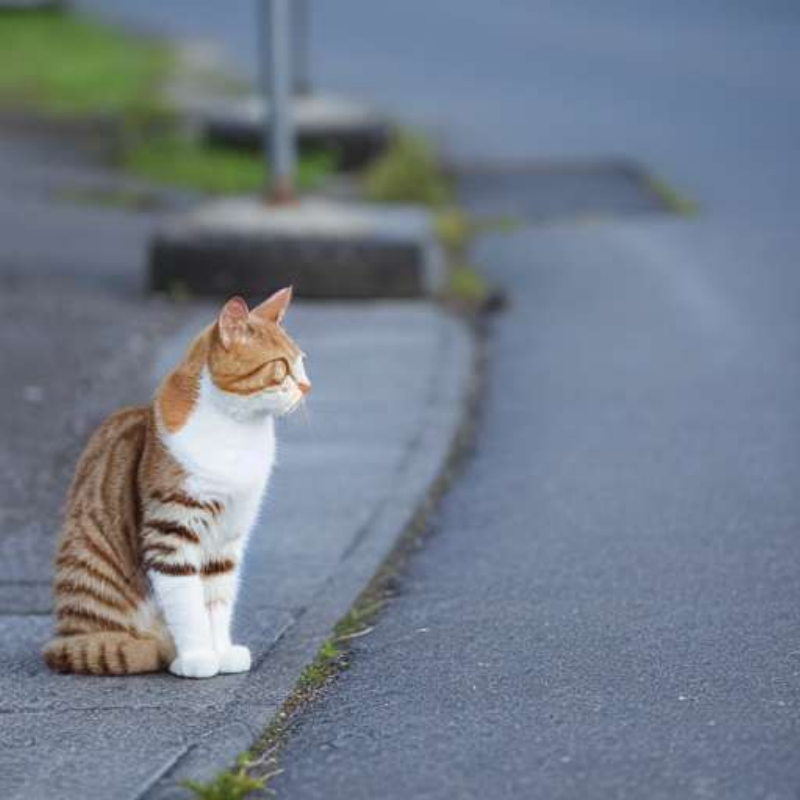}
  }
  \vspace{-12pt}
  \subfloat[(e) Input]
  {
      \includegraphics[width=0.32\linewidth]{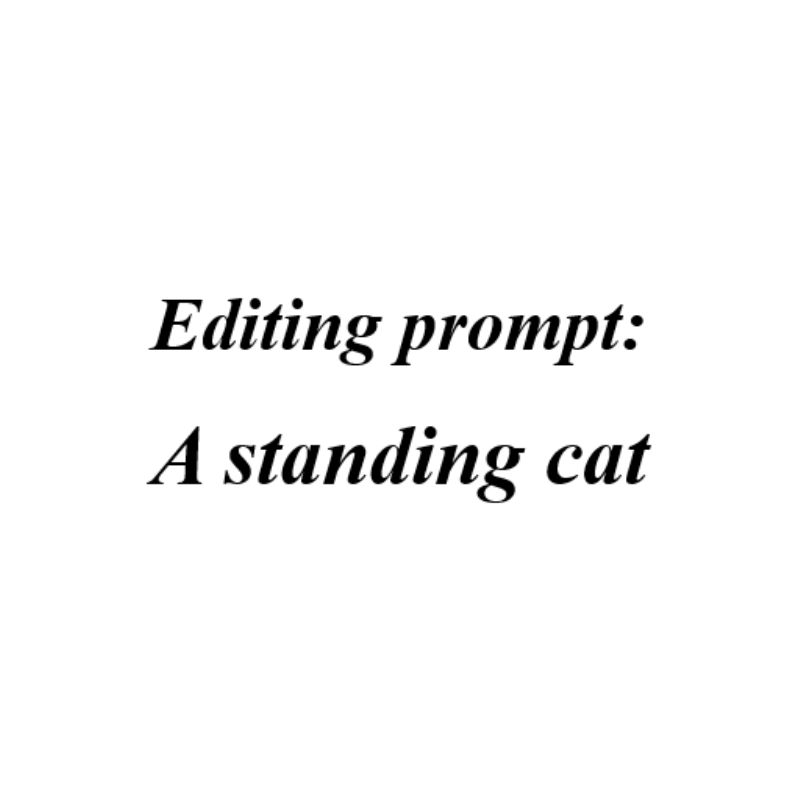}
  }
  \vdashlinedown
  \subfloat[(h) MasaCtrl]
  {
      \includegraphics[width=0.32\linewidth]{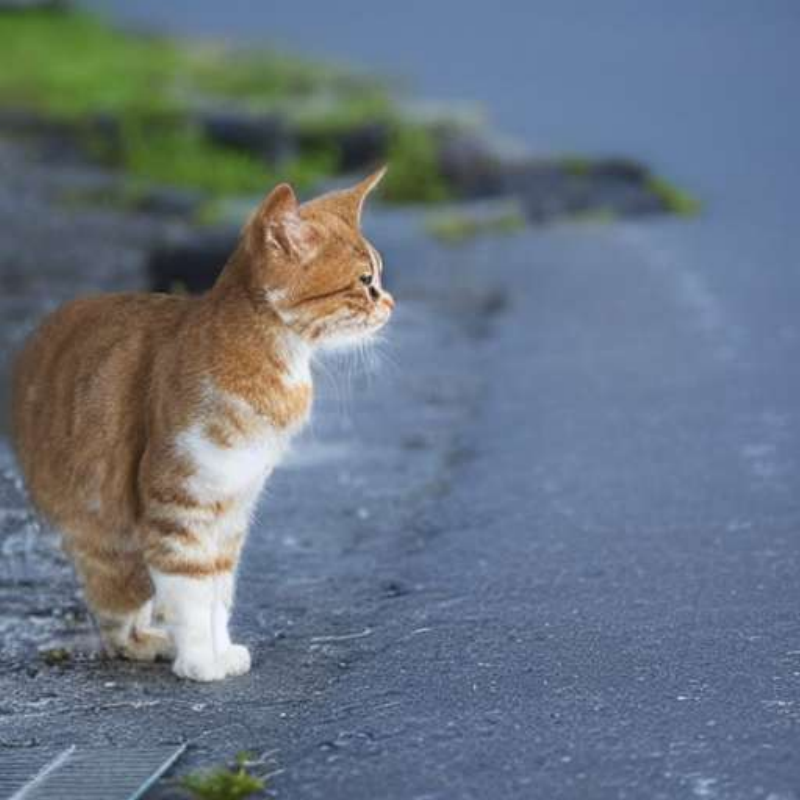}
  }
  \subfloat[(i) ~~Move\&Act\newline\centering(Ours)]
  {
      \includegraphics[width=0.32\linewidth]{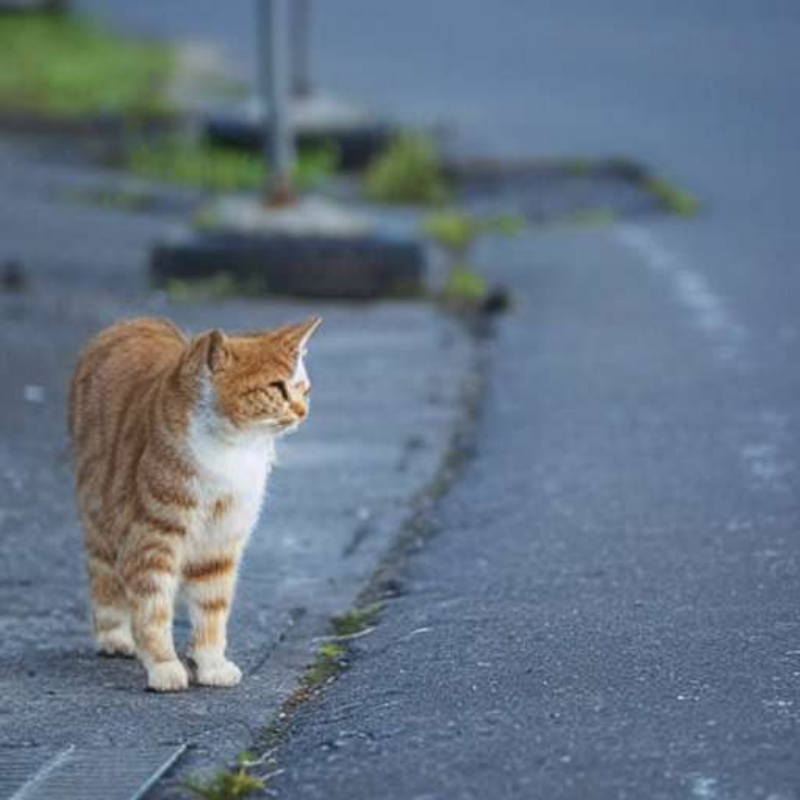}
  }  
  \caption{(a)\,-\,(d): our method allows users to control where to place the edited object. (e)\,-\,(i): altering only action, our method captures the ``standing'' pose and maintains consistent backgrounds with the input image.}
  \label{fig1}
\end{figure}

\section{Introduction}
\label{sec:intro}

Recent advancements in text-to-image generation using diffusion models~\cite{ref2, ref3, ref4, ref5} have significantly improved image generation capabilities. By capitalizing on extensive training data, these models~\cite{ref3, ref4} excel in generating remarkable images based on user prompts. Compared to generative adversarial networks (GANs), diffusion models provide superior quality and diversity in image generation. Consequently, numerous studies~\cite{ref6, ref7, ref8, ref9, ref10, ref11} have begun investigating diffusion models for image editing tasks. Predominantly, these methods involve text-conditioned editing, where users supply text input to modify images. Although techniques like~\cite{ref6, ref7, ref8, ref9} can adeptly edit image style, object appearance, and object category based on text input, they do not alter scene layout or modify object actions in the image. Nonetheless, there are instances where users desire to edit an object's action while maintaining its appearance and background. This specific editing process is called consistent image editing.

\begin{figure*}[!t]
  \centering
  \includegraphics[width=1.0\textwidth]{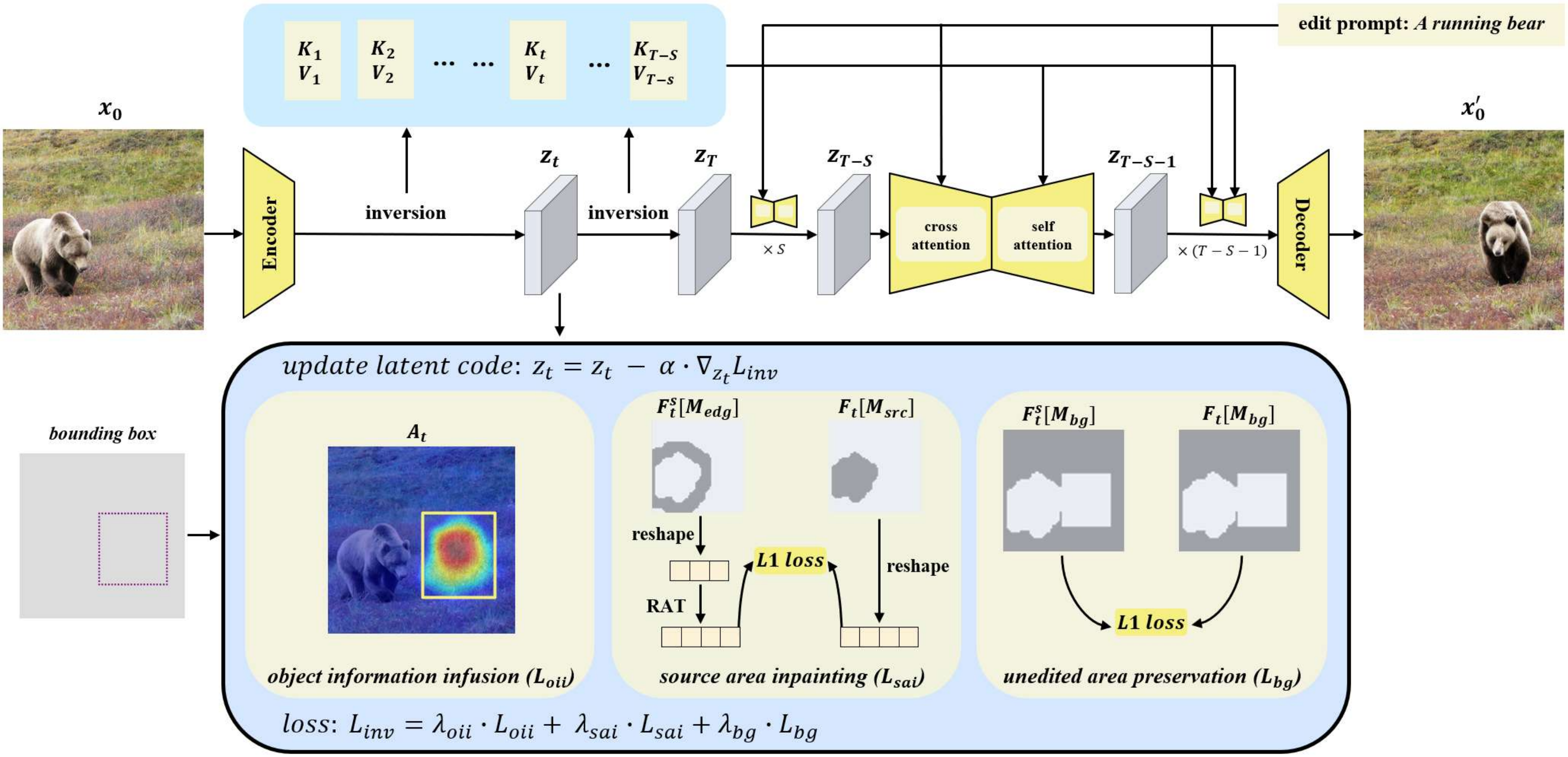}
  \caption{Move\&Act pipeline. In the inversion stage, we update latent code and transfer object information to the target area while repairing the source background. In the reverse stage, we use self-attention image features for consistent image editing.}
  \label{fig2}
\end{figure*}

Imagic~\cite{ref40} adopts a complex prompt fine-tuning process for text-based consistent image editing. MasaCtrl~\cite{ref10} offers a tuning-free approach, transforming self-attention into mutual self-attention. These methods edit an object's action without allowing control over the object's generated position. DragonDiffusion~\cite{ref21} facilitates object displacement within the image, yet it restricts the editing of the relocated object's action. In practical applications, users may require simultaneous editing of an object's action and position. As illustrated in the upper part of Figure\,\ref{fig1}, we allow users to input editing text to adjust the object's action, while also providing a bounding box to dictate the object's generated position. 

We also maintain better background preservation. By designating the object's generation position as its original location, our method effectively reduces to MasaCtrl's editing function, solely modifying the object's action. In the lower part of Figure\,\ref{fig1}, the backgrounds generated by InstructPix2Pix~\cite{ref42}, PnP~\cite{ref7} and MasaCtrl~\cite{ref10} exhibit a certain loss in detail, whereas our method produces a background more closely resembling the input image. Also, the cat's action in our results aligns more accurately with the editing prompt and the surrounding environment.

We introduce a training-free Move\&Act method, for consistent image editing, empowering users to simultaneously modify an object's action and its generated position within an image. As depicted in Figure\,\ref{fig2}, our method transfers object information from the original area to a user-defined target bounding box, effectively creating the edited object in the desired location. Since cross-attention maps linked to an object token encompass spatial position information of the object in the image, we propose the transfer of high-response areas from the original region to the target region.

To achieve this, we maximize attention scores in the target area while minimizing them in non-target regions. However, this strategy alone is inadequate for a comprehensive object information transfer, as larger attention scores may linger in the original area. To address this, we utilize the surrounding background to inpaint the object's initial location, employing an efficient feature extractor to obtain features from the original and surrounding background regions, and calculating the $L_1$ loss between them. This inpainting process not only harmonizes the background in the original and adjacent areas but also effectively eliminates residual object information, facilitating a complete transfer to the target region. Moreover, we incorporate a background preservation loss to guarantee consistency before and after editing. The object information transfer procedure is finalized during the inversion stage, while the subsequent reverse stage ensures consistency in the object's appearance and image background by leveraging the image feature in self-attention to query the key and value at the corresponding inversion stage timestep, as opposed to the image feature itself.

We have performed extensive experiments to substantiate our efficacy. Both qualitative and quantitative outcomes demonstrate that our Move\&Act proficiently edits the object's action in line with the editing prompt and accurately generates the edited object within the target bounding box.

\section{Related work}

\subsection{Text-to-Image Generation}

Recent research utilizing diffusion models have made significant strides in the realm of image generation. Notably,~\cite{ref1} has outperformed the majority of generative adversarial networks (GANs) in terms of generation metrics, indicating a substantial progression in the field. The use of diffusion models in the generation of images from text has produced remarkable outcomes. 

To align generated images and prompts, Glide~\cite{ref2} integrated diffusion models with guidance techniques. Large-scale diffusion models have been created~\cite{ref3, ref4}, exhibiting even more remarkable performance in generating images from text. However, the substantial size of these models renders them unsuitable for average users with limited computing capabilities. \cite{ref5} proposed the latent diffusion model (LDM), which employs an autoencoder to map images from the image space to the latent space, thereby reducing the computational demands of diffusion models. Furthermore, LDM incorporates a cross-attention layer within its structure, achieving outstanding results in the text-to-image task. Our work builds on LDM, specifically utilizing a pretrained LDM called stable diffusion~\cite{ref5}.

\subsection{Image Editing}

\subsubsection{Text Driven Image Editing.}
Recently, diffusion model-based image editing has achieved unparalleled results. For example, P2P~\cite{ref6} utilizes cross-attention control to edit images from one prompt to another, while PnP~\cite{ref7} introduces a plug-and-play method for text-driven image translation. Pix2Pix-Zero~\cite{ref8} utilizes regularized forward DDIM and cross-attention guidance to achieve text-driven image translation. PTI~\cite{ref9} edits particular objects by optimizing the condition embedding during the image reconstruction process. This updated condition embedding is then utilized to finalize the image editing. 
NTI~\cite{ref41} discovers that optimizing null-text Embedding can accurately reconstruct the original image while avoiding fine-tuning the model, thereby preserving the structure of image and enabling high-fidelity editing of real images.

\subsubsection{Consistent Image Editing.}
Consistent image editing~\cite{ref40, ref10} seeks to modify a single object in an image while leaving the rest of the image intact. DragGAN~\cite{ref12} was the first to introduce drag-style image editing to achieve this consistency. However, the quality and generalizability of DragGAN's editing are constrained by its dependence on GANs. Capitalizing on the success of diffusion models,~\cite{ref21, ref22} have employed diffusion models to execute drag-style image editing, resulting in promising outcomes.

Imagic~\cite{ref40} necessitates optimizing both textual embedding and diffusion model. It linearly interpolates the target text embedding and the optimized embedding, and then inputs the fine-tuned model to generate the edited image. MasaCtrl~\cite{ref10} employs mutual self-attention which involves querying image content from source images using a self-attention mechanism to achieve consistent image editing. These methods are restricted to modifying the actions of objects and lack control over the generated position. In contrast, ours consists of an inversion branch and an edit branch, allowing for simultaneous editing of the object's action and control over its generated position.

\section{Methodology}
The pipeline of our Move\&Act is depicted in Figure\,\ref{fig2}, where we are given an image and an object to be edited. Our objective is to generate the edited object within the designated target bounding box while ensuring that it adheres to the editing prompt. In the subsequent sections, we refer to the user-provided bounding box area as the target area, the region where the object originally resides as the source area, and the complement of the union of both target and source areas as the unedited area (or background area). 

To accomplish this editing task, we employ a two-step process. Firstly, we transfer the object information from the source area to the target area during a specific step in the inversion stage. Secondly, we modify the object's action throughout the reverse stage, resulting in a seamlessly edited image with a sophisticated and coherent appearance.

\subsection{Object Information Transfer}
To facilitate the generation of the edited object within the target area, we infuse the object information into this designated area by updating the latent code during the 35-$th$ time step of the inversion process. Following the relocation of the object to the target area, it is essential to seamlessly inpaint the source area. In the ensuing sections, we will delve deeper into the mechanics of object information infusion and the intricacies of source area inpainting.

\subsubsection{Object Information Infusion.}
Our observation reveals that the attention map, associated with each object token within the cross-attention mechanism of stable diffusion, encompasses spatial position information about the object within the image. This observation aligns with the findings of previous studies~\cite{ref18, ref19}. Leveraging this insight, several studies~\cite{ref19, ref37, ref38, ref39} have utilized this mechanism to manipulate the arrangement of objects within images. In a similar vein to~\cite{ref18, ref19}, we utilize a 16$\times$16 resolution attention map to update the latent code, as this particular resolution attention map is rich in semantic information. Consequently, to facilitate the relocation of the object to the target area, we ensure that the attention map, corresponding to the edited object, achieves the highest attention score within the confines of the target area:
\begin{equation}
  L_{in} = 1 - \frac{1}{k}\Sigma top(A_t[M_{tgt}], k),
\end{equation}
where $M_{tgt}$ symbolizes the mask of the target area, and $A_{t}$ represents the cross attention map that corresponds to the $t$-$th$ time step of the edited object token, with $t=35$. The $top(\cdot, k)$ operation is employed to extract the largest $k$ attention scores within the target area.

Conversely, we minimize the attention score attributed to areas outside the designated target area, ensuring a precise and controlled editing process:
\begin{equation}
  L_{out} = \frac{1}{N}\Sigma A_t[1-M_{tgt}].
\end{equation}

The final loss for object information infusion becomes:
\begin{equation}
  L_{oii} = L_{in} + L_{out}.
\end{equation}

\begin{figure}[!t]
  \centering
  \captionsetup[subfloat]{labelsep=none,format=plain,labelformat=empty}
  \subfloat[(a) Input]   
  {
      \includegraphics[width=0.23\linewidth]{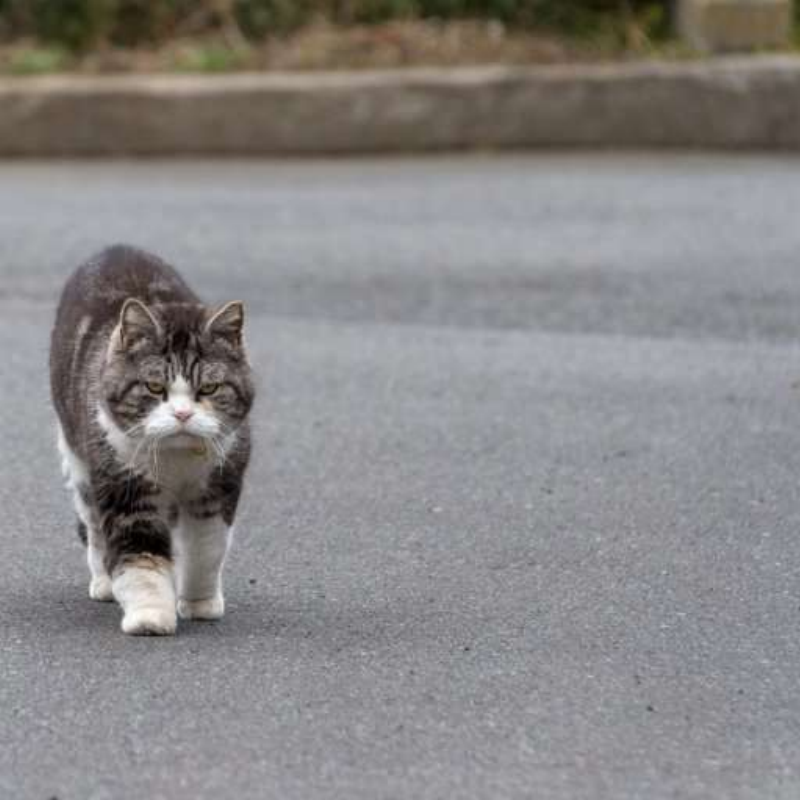}
  }
  \subfloat[(b) CA map]
  {
      \includegraphics[width=0.23\linewidth]{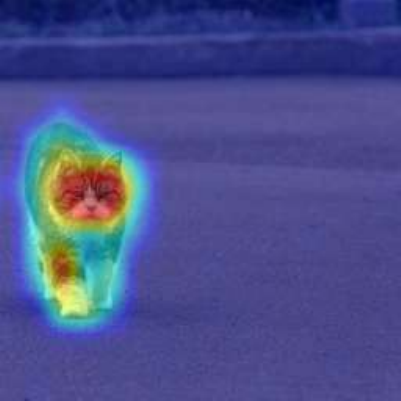}
  }
  \subfloat[(c) Source area]
  {
      \includegraphics[width=0.23\linewidth]{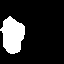}
  }
  \subfloat[(d) Edge area]
  {
      \includegraphics[width=0.23\linewidth]{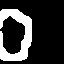}
  }
  \caption{We use cross attention map to locate the source area of the object, and use dilation operation to obtain the edge area around the source region.}
  \label{fig3}
\end{figure}

\subsubsection{Source Area Inpainting.}
To perform inpainting on the source area, we must first accurately locate it. Given that the high-response region of the cross-attention map corresponding to the object token represents the object's spatial position within the image, we employ a threshold to isolate this high-response area, which serves as the object's source area, as illustrated in Figure\,\ref{fig3}. To circumvent conflicts arising from maximizing the attention score of the target area, we inpaint the actual source area by subtracting the union of the source and target areas from the original source area.

Intuitively, the background of the source area should exhibit similarities to the surrounding background. To achieve this, we first conduct a dilation operation on the mask corresponding to the source area, thereby obtaining the region surrounding the source area, as demonstrated in Figure\,\ref{fig3}. We denote the expanded surrounding area as the edge area.

In order to harmonize the background of the source area with that of the edge area, we extract features from both areas of the latent code and subsequently compute the $L_1$ loss between the features of these regions. DIFT~\cite{ref20} discovered that the UNet of the diffusion model is an efficient feature extractor, thus we utilize UNet to extract features from the latent code. The decoder of the UNet denoiser comprises four blocks of varying scales. According to DIFT, the second layer contains more semantic information, while the third layer encompasses more geometric information. In our method, we opt for the third layer.

Since the dimensions of the edge area and the source area may not necessarily be equal, we repeat and truncate the features of the edge area, to align features in both regions. The loss for source area inpainting becomes:
\begin{equation}
  L_{sai} = \Vert F_t[M_{src}] - RAT(F_t^s[M_{edg}]) \Vert_1,
\end{equation}
where $M_{src}$ and $M_{edg}$ represent the source area and edge area. $F_t^s$ denotes the feature of the original latent code, and $F_t$ symbolizes the feature of the updated latent code, with $t=35$. $RAT(\cdot)$ is a repeat and truncate operation.

\begin{figure}[!t]
  \centering
  \captionsetup[subfloat]{labelsep=none,format=plain,labelformat=empty}
  \rotatebox{90}{\scriptsize{~~~~~~w/o $L_{sai}$}}
  \hspace{-3pt}
  \subfloat   
  {
      \includegraphics[width=0.22\linewidth]{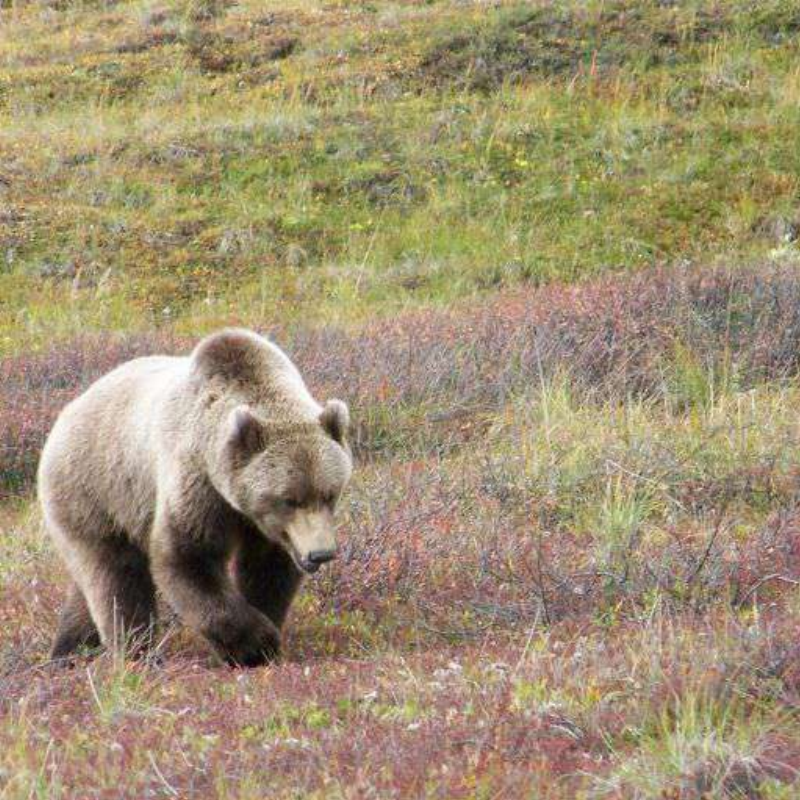}
  }
  \subfloat
  {
      \includegraphics[width=0.22\linewidth]{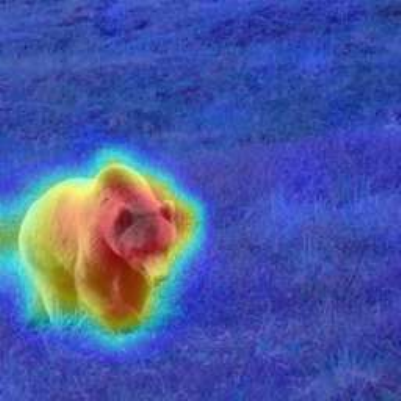}
  }
  \subfloat
  {
      \includegraphics[width=0.22\linewidth]{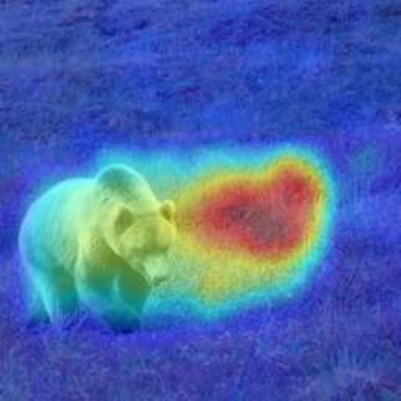}
  }
  \subfloat
  {
      \includegraphics[width=0.22\linewidth]{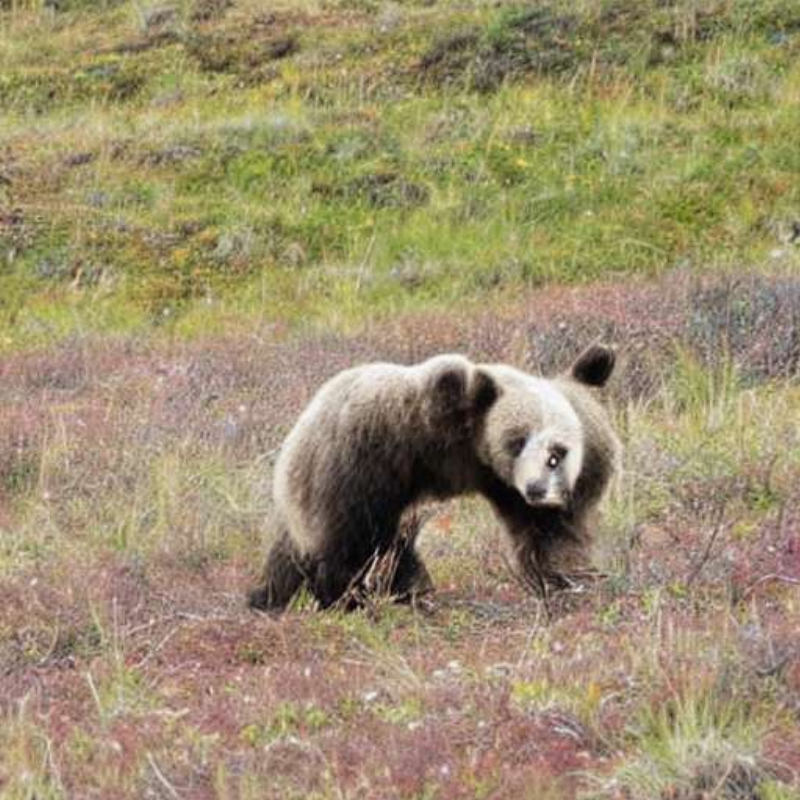}
  }

  \rotatebox{90}{\scriptsize{~~~~~~~w/ $L_{sai}$}}
  \hspace{-3pt}
  \subfloat[(a) Input]   
  {
      \includegraphics[width=0.22\linewidth]{fig4/0096.pdf}
  }
  \subfloat[(b) Before]
  {
      \includegraphics[width=0.22\linewidth]{fig4/before.pdf}
  }
  \subfloat[(c) After]
  {
      \includegraphics[width=0.22\linewidth]{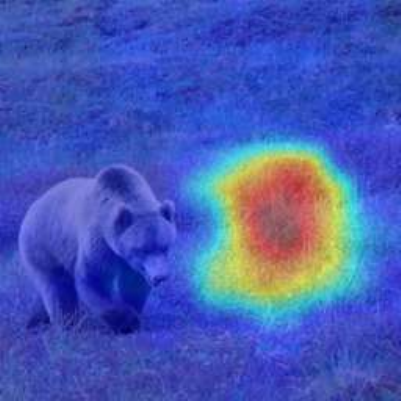}
  }
  \subfloat[(d) Output]
  {
      \includegraphics[width=0.22\linewidth]{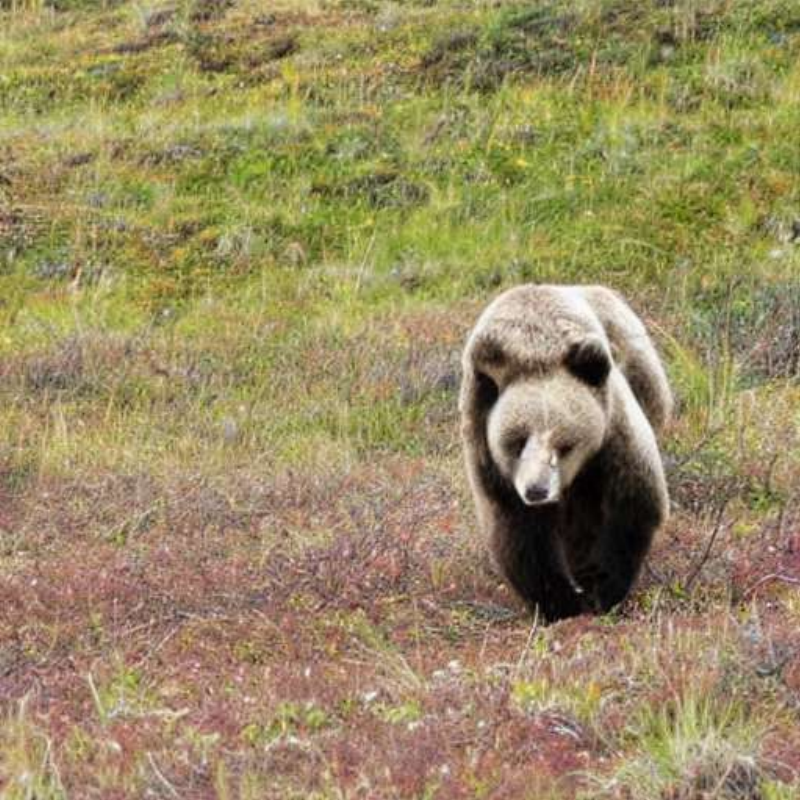}
  }
  \caption{Efficacy of source area inpainting loss $L_{sai}$. With/Without $L_{sai}$, the object fully/partly transfers to target area.}
  \label{fig4}
\end{figure}

Apart from its impact on the source area, $L_{sai}$ also serves a potentially crucial role. Employing solely $L_{oii}$ may not facilitate the complete transfer of object information to the target area, even though $L_{oii}$ minimizes the attention scores of the non-target area. In Figure\,\ref{fig4}, when only $L_{oii}$ is utilized, remnants of the object information persist in the source area, which could hinder the generation of the edited object within the target area. By concurrently employing $L_{sai}$ and $L_{oii}$, $L_{sai}$ will completely erase the object in the source area.

\subsection{Background Preservation}
Upon relocating the edited object from the source area to the target region, we observe that the infusion of object information and the subsequent inpainting of source area may inadvertently cause damage to the image's background. To address this issue and maintain the integrity of the unedited areas, we propose a background-preserving loss function.

\begin{figure}[!t]
  \centering
  \subfloat[Input]
  {
      \includegraphics[width=0.23\linewidth]{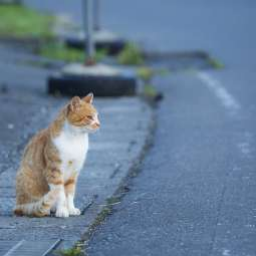}
  }
  \subfloat[MasaCtrl]
  {
      \includegraphics[width=0.23\linewidth]{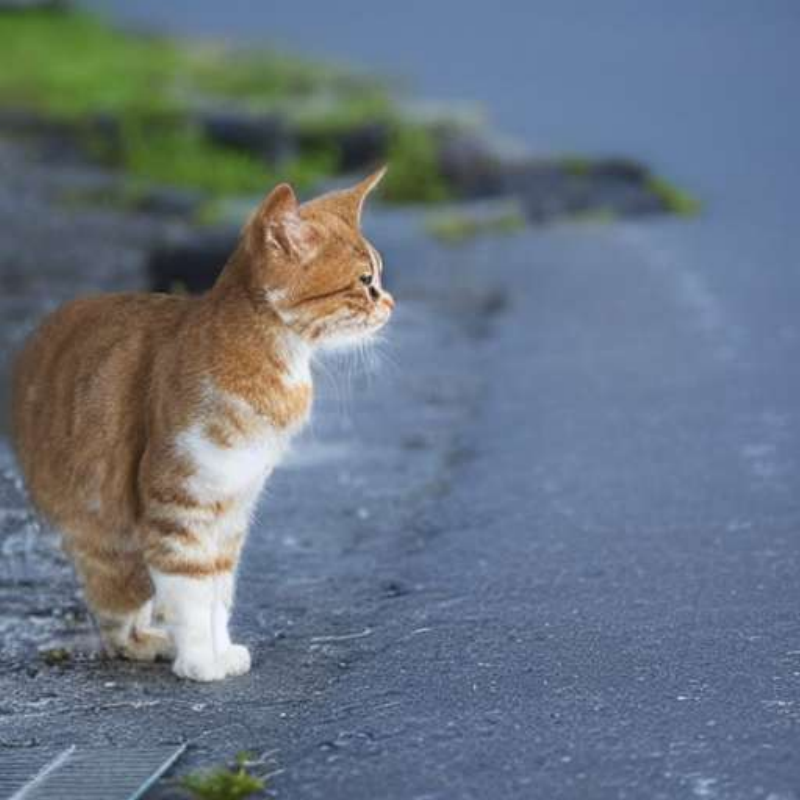}
  }
  \subfloat[w/o $L_{bg}$]
  {
      \includegraphics[width=0.23\linewidth]{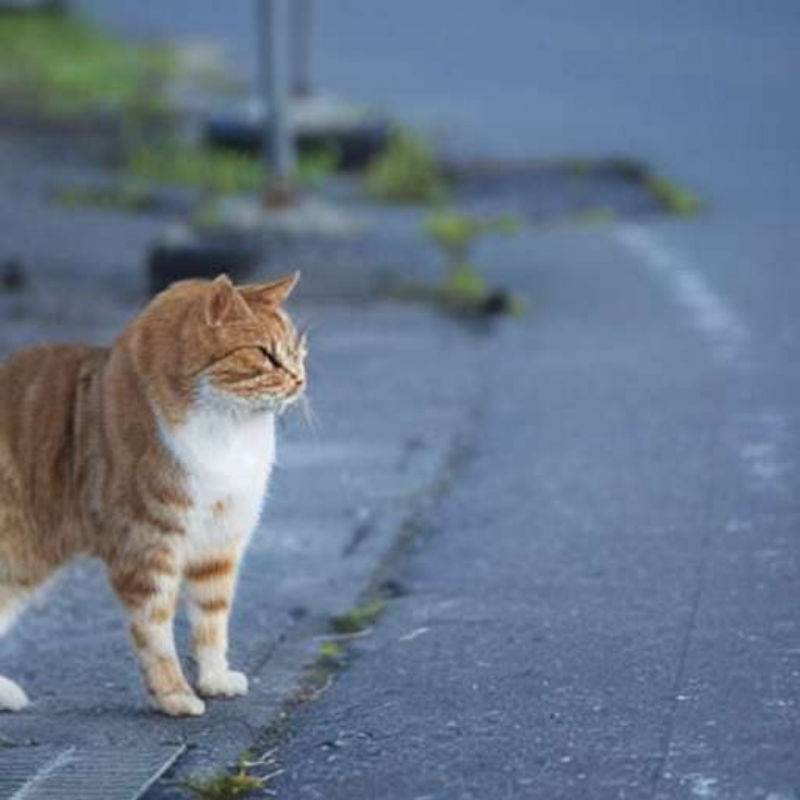}
  }
  \subfloat[w/ $L_{bg}$]
  {
      \includegraphics[width=0.23\linewidth]{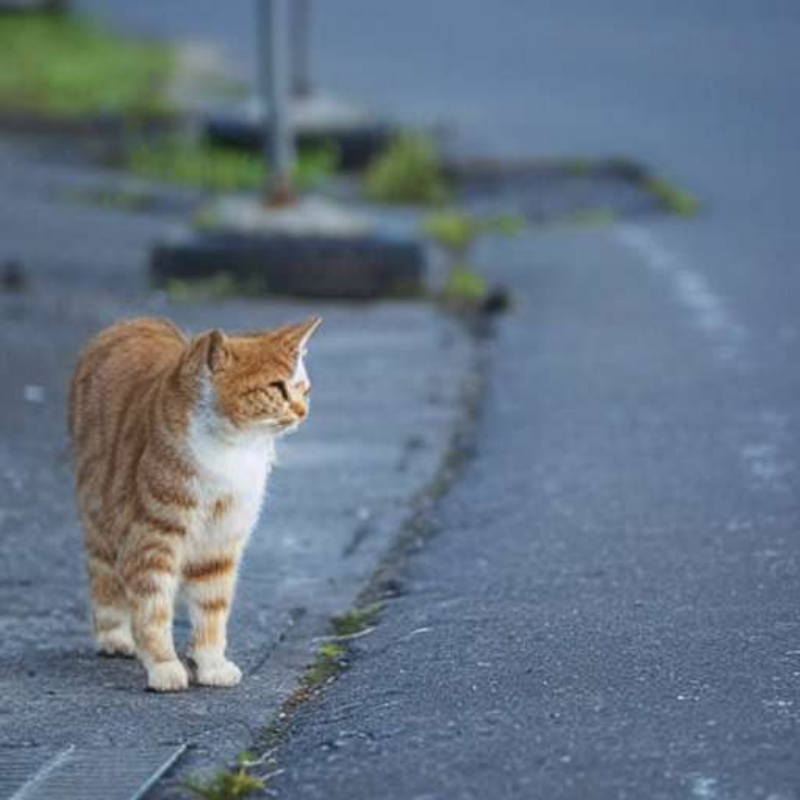}
  }
  \caption{Editing prompt: A standing cat. Efficacy of background preservation loss $L_{bg}$. Without $L_{bg}$, background details like meadow become distorted resembling MasaCtrl. With $L_{bg}$, the background remains well-preserved.}
  \label{fig5}
\end{figure}

Utilizing the rich image structure and texture information contained within the intermediate features of the UNet denoiser, we employ UNet to extract features from the latent code. Subsequently, we compute the $L_1$ loss for the unedited regions between the original and updated latent codes:
\begin{equation}
  L_{bg} = \Vert F_t[M_{bg}] - F_t^s[M_{bg}] \Vert_1,
\end{equation}
where $M_{bg}$ represents the unedited area. As illustrated in Figure\,\ref{fig5}, the absence of $L_{bg}$ may result in the loss of some background information in the edited image. However, by introducing $L_{bg}$, we ensure consistency between the pre- and post-editing background areas. Simultaneously, the object information in the unedited region is suppressed, guaranteeing that the edited object adheres more closely to the user-specified bounding box constraints.

\begin{figure*}[!t]
  \centering
  \captionsetup[subfloat]{labelsep=none,format=plain,labelformat=empty}
  
  \subfloat[(a.1) Input]
  {
      \label{fig:subfig6}\includegraphics[width=0.15\textwidth]{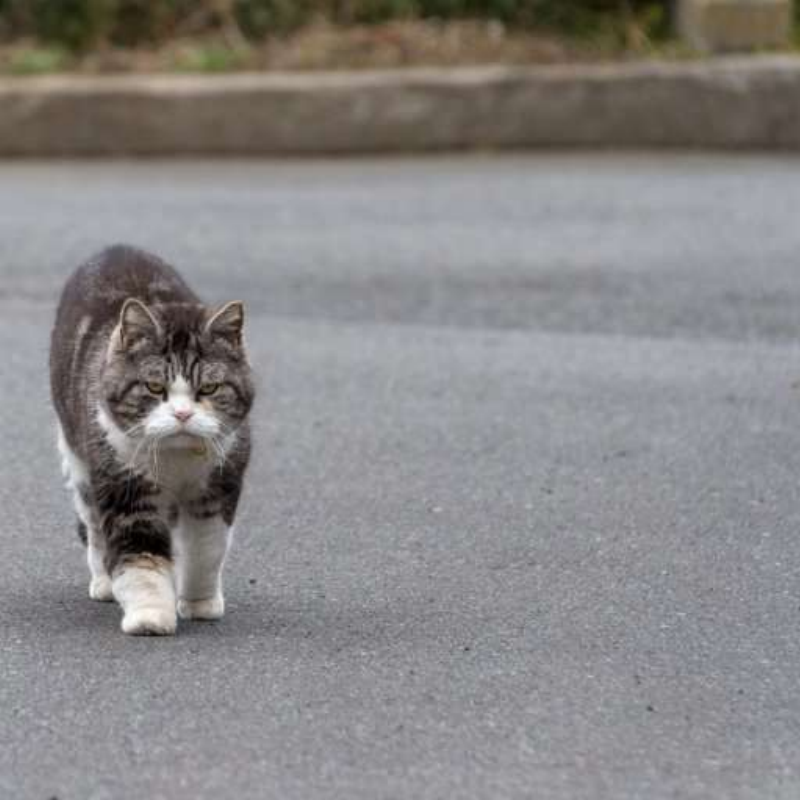}
  }
  \subfloat[~(a.2) \textit{A walking cat}]   
  {
      \label{fig:subfig6-bbox}\includegraphics[width=0.15\textwidth]{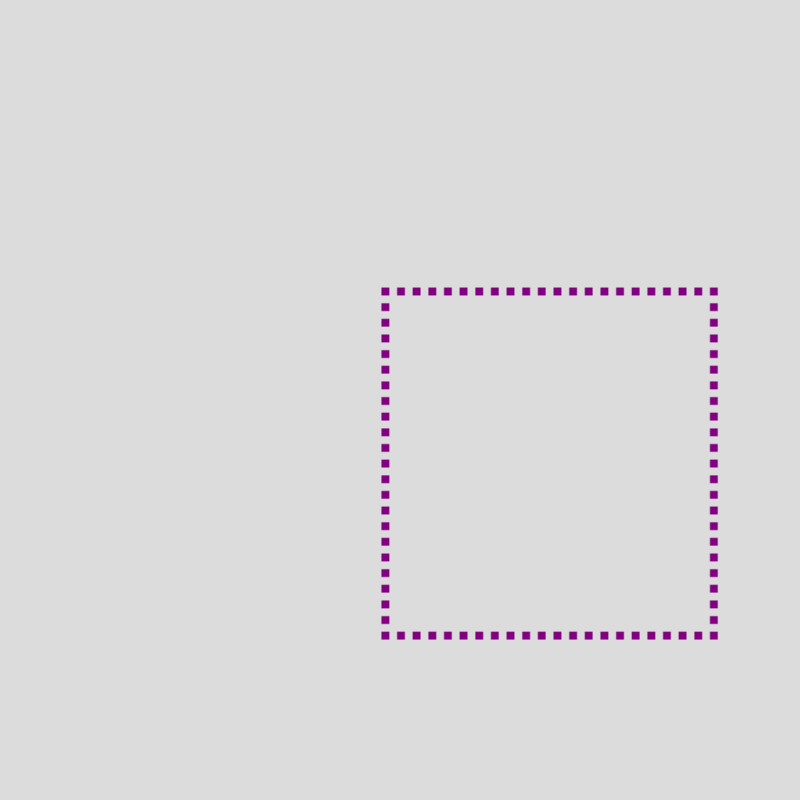}
      \label{fig:subfig6-output}\includegraphics[width=0.15\textwidth]{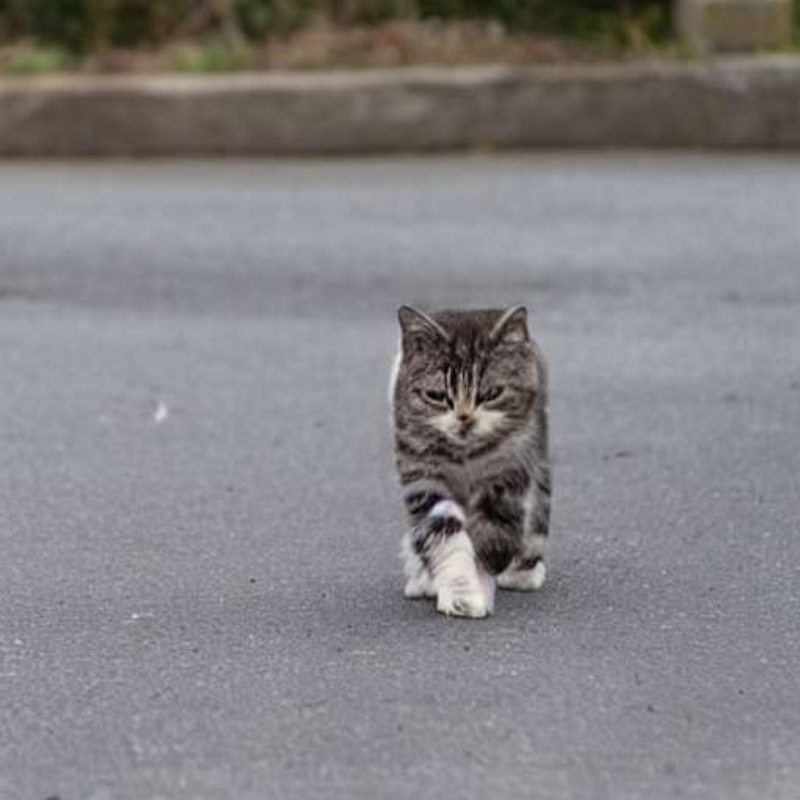}
  }
  \subfloat[(b.1) Input]
  {
      \includegraphics[width=0.15\textwidth]{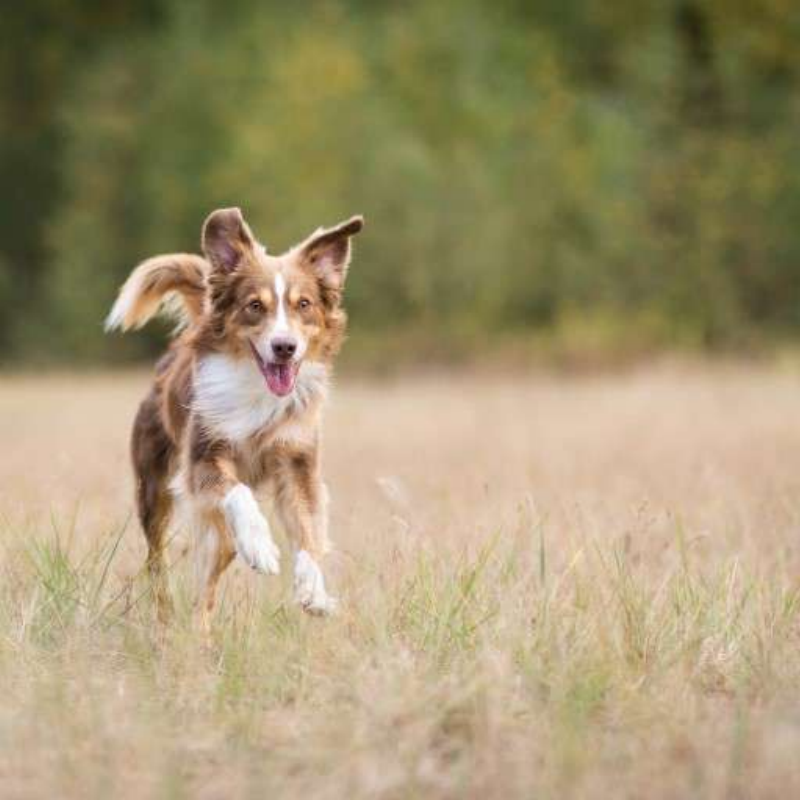}
  }
  \subfloat[~(b.2) \textit{A sitting dog}]   
  {
      \includegraphics[width=0.15\textwidth]{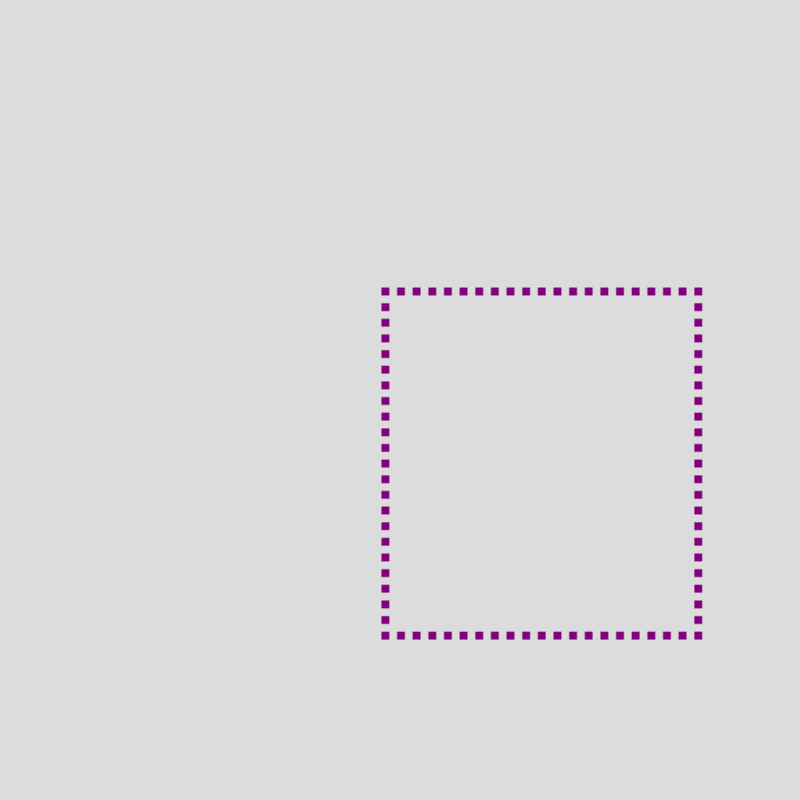}
      \includegraphics[width=0.15\textwidth]{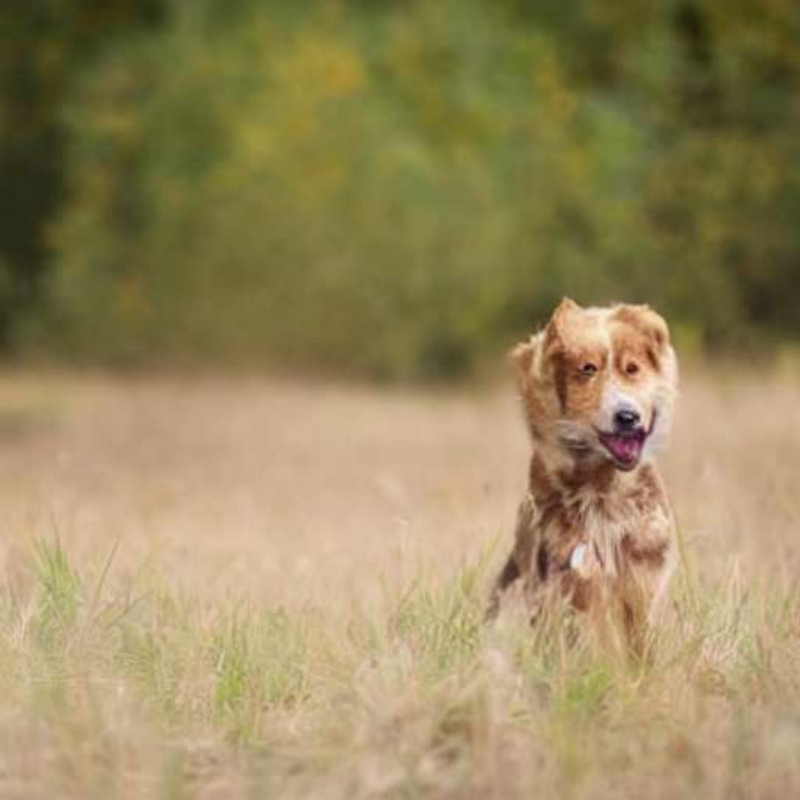}
  }

  
  \subfloat[(c.1) Input]
  {
      \includegraphics[width=0.15\textwidth]{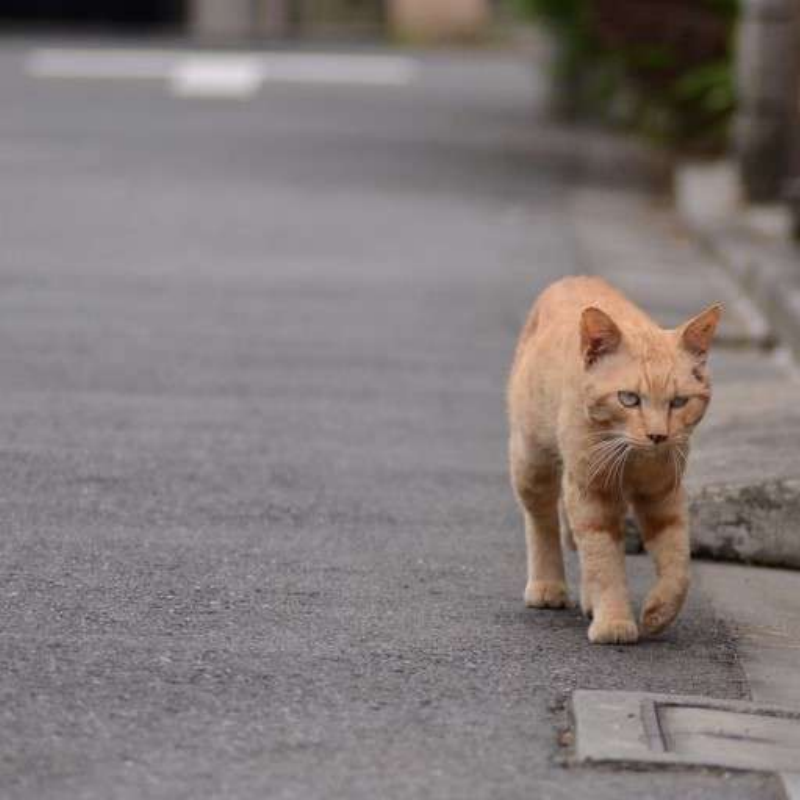}
  }
  \subfloat[~(c.2) \textit{A running cat}]   
  {
      \includegraphics[width=0.15\textwidth]{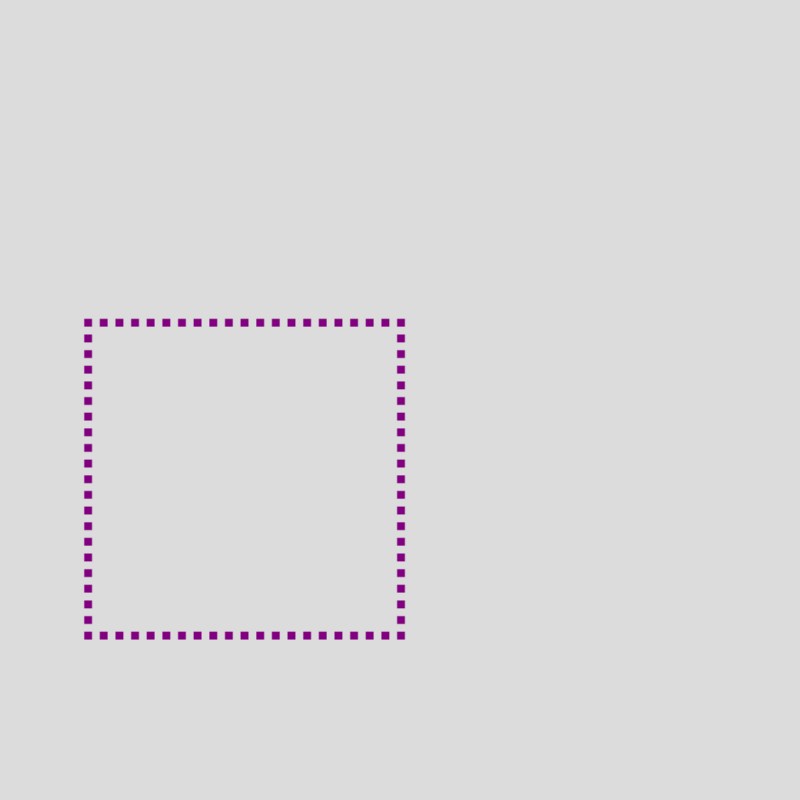}
      \includegraphics[width=0.15\textwidth]{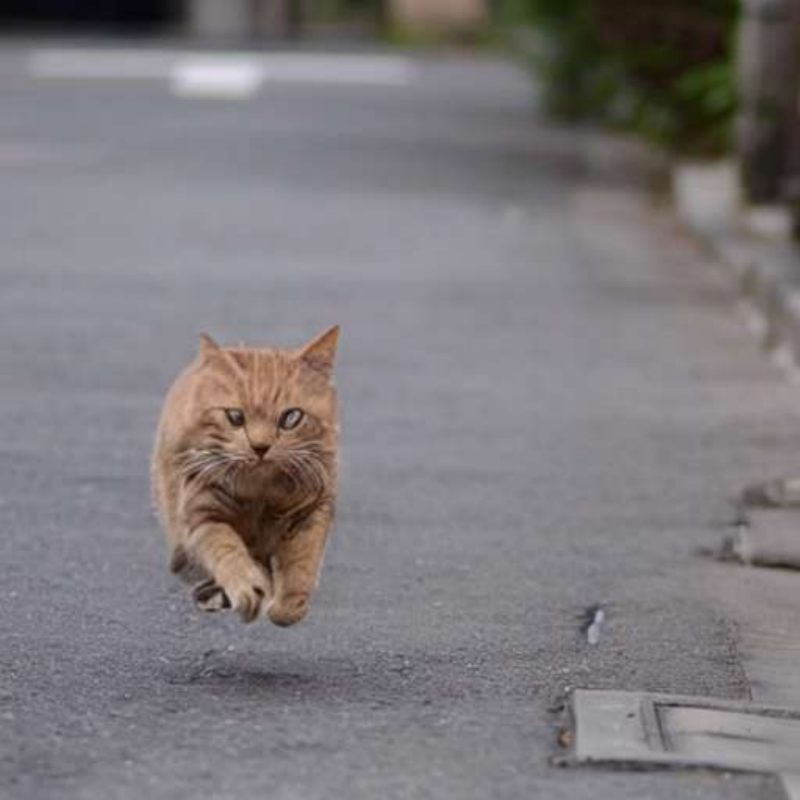}
  }
  \subfloat[(d.1) Input]
  {
      \includegraphics[width=0.15\textwidth]{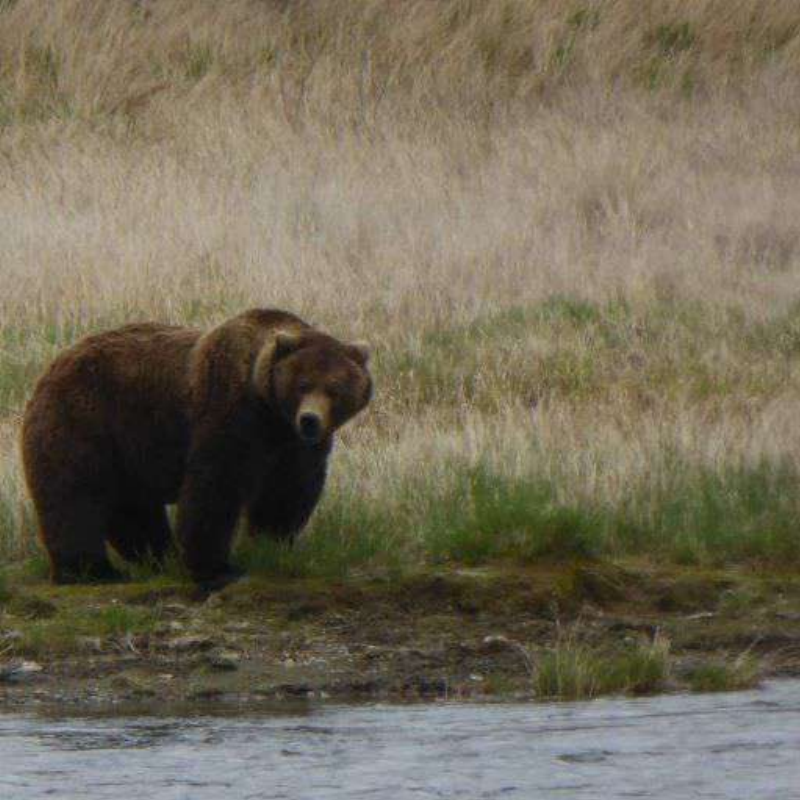}
  }
  \subfloat[~(d.2) \textit{A walking bear}]   
  {
      \includegraphics[width=0.15\textwidth]{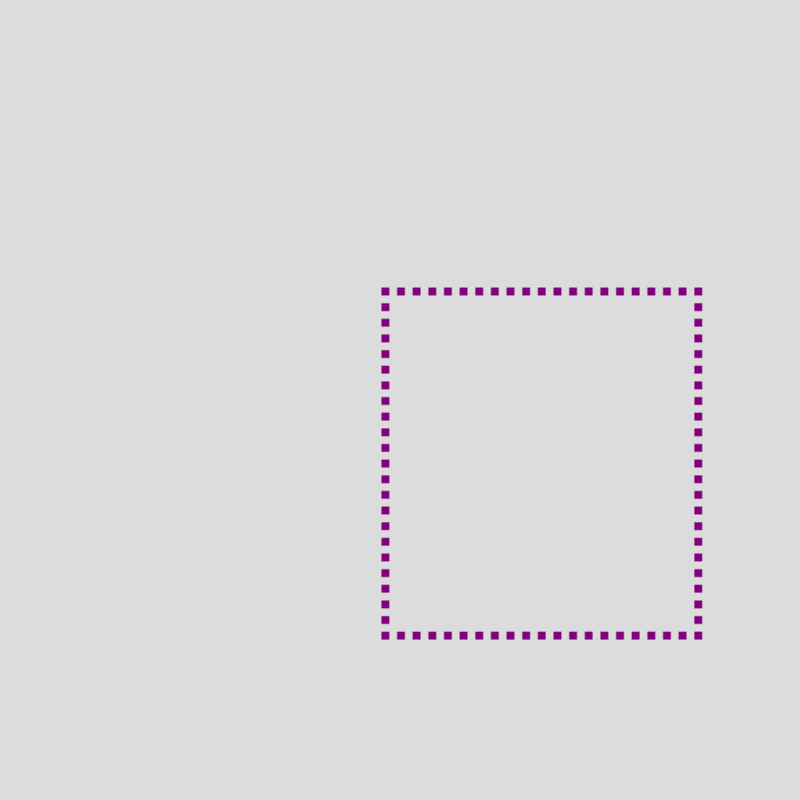}
      \includegraphics[width=0.15\textwidth]{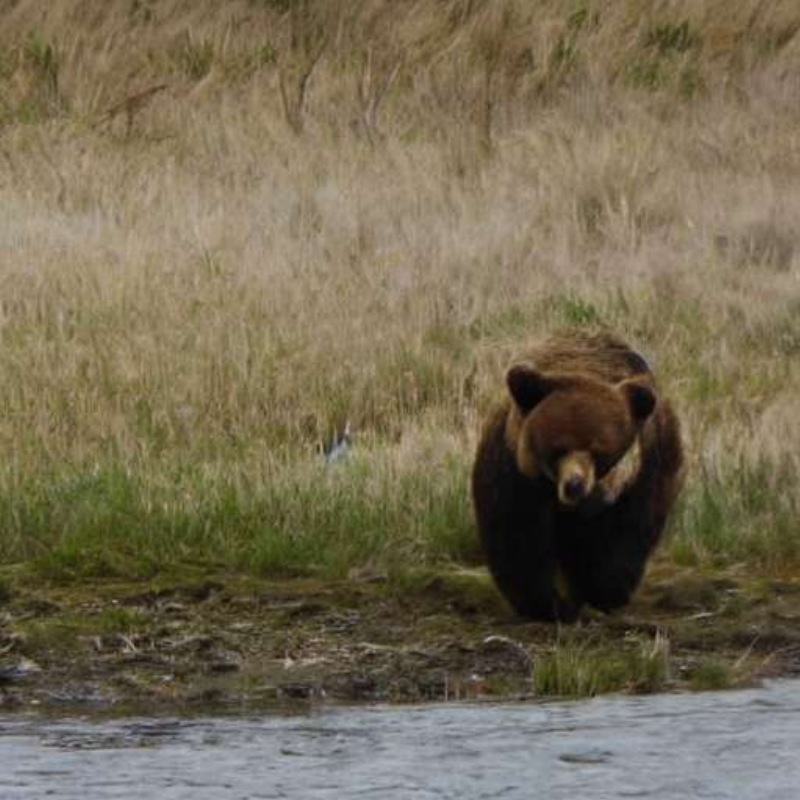}
  }


  \caption{Visualization of our proposed Move\&Act for consistent image editing.}
  \label{fig6}
\end{figure*}

The ultimate loss function utilized for infusing object information and preserving background integrity at the 35-$th$ time step of inversion process is:
\begin{equation}\label{eq:inversion-loss}
  L_{inv} = \lambda_{oii} \cdot L_{oii} + \lambda_{sai} \cdot L_{sai} + \lambda_{bg} \cdot L_{bg},
\end{equation}
where $\lambda_{oii}$, $\lambda_{sai}$, and $\lambda_{bg}$ serve as balancing parameters. By applying $L_{inv}$, latent code $z_{t}$ is updated with a step of $\alpha$:
\begin{equation}\label{eq:update}
  z_{t} = z_{t} - \alpha \cdot \nabla_{z_{t}} L_{inv}.
\end{equation}

As updating step increases, the step size $\alpha$ undergoes linear decay. This approach ensures a smooth and consistent image editing process, preserving the background and seamlessly integrating the edited object into the target area.

\subsection{Content Editing Consistency}
Upon obtaining $z_{T}$ through the inversion process, we employ it as the initial point for the reverse denoising procedure. To ensure that the edited object generates the same action as indicated by the editing prompt, we integrate the editing prompt into the reverse process using the cross attention mechanism of UNet. Besides, to preserve the consistency of the image content prior to and following the editing process, we replace the key and value within the self attention mechanism with their counterparts from the corresponding time step in the inversion stage, commencing from the 7-$th$ time step during the reverse procedure.

In contrast to the approach presented in~\cite{ref10}, we query the key and value during the inversion stage, as we believe that the key and value from this stage contain richer information about the original image's object appearance and background compared to using the key and value from the reconstruction stage as in~\cite{ref10}. Furthermore, utilizing the key and value from the inversion stage diminishes the image reconstruction cost. Consequently, our optimization process transforms the original three branches of inversion, reconstruction, and editing~\cite{ref10} into two branches: inversion and editing. The exchange of key and value in the optimized operation can be expressed:
\begin{equation}\label{eq:kv-exchange}
    f = 
    \left\{ 
        \begin{array}{lc}
            attention(Q, K_{inv}, V_{inv}), \; & t > S \; and \; l \in  L, \\
            attention(Q, K, V), & otherwise, \\
        \end{array}
    \right.
\end{equation}
where $K_{inv}$ and $V_{inv}$ denote the key and value of the corresponding time step in the inversion stage. When the time step exceeds $S = 7$ and the self attention layer is part of the predefined set $L$, we alter the self attention key and value to $K_{inv}$ and $V_{inv}$. By enabling image features to access and interact with the key and value of the input image, we facilitate a uniform and congruent image editing process, ultimately yielding a cohesive and polished result.

\section{Experiments}

\begin{figure*}[!t]
  \centering
  \captionsetup[subfloat]{labelsep=none,format=plain,labelformat=empty}
  \rotatebox{90}{\normalsize{~~~~~~\textit{A running dog}}}
  \hspace{-3pt}
  \subfloat
  {
      \includegraphics[width=0.18\textwidth]{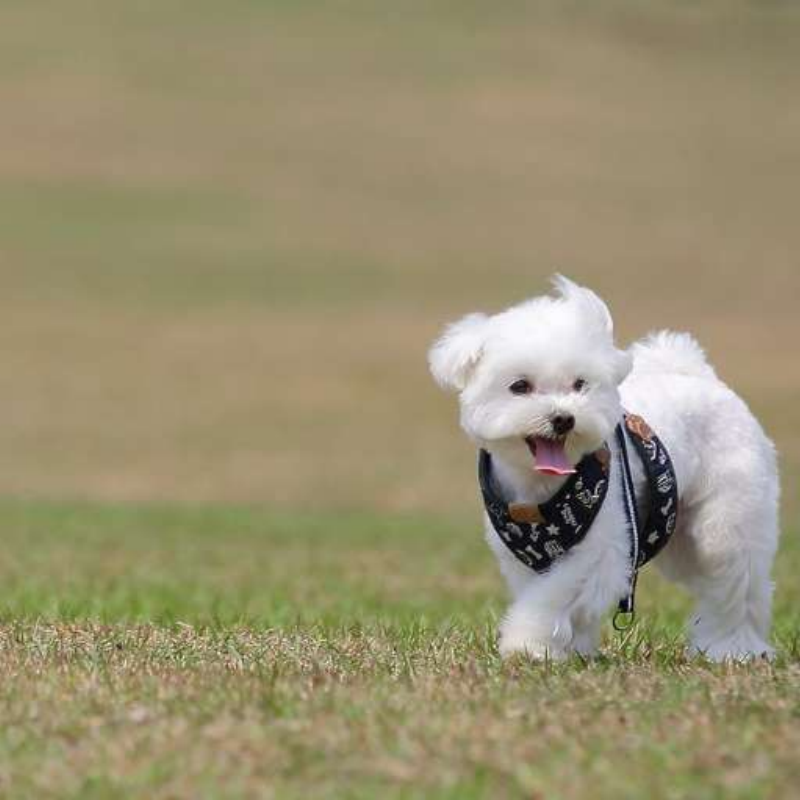}
  }
  \subfloat
  {
      \includegraphics[width=0.18\textwidth]{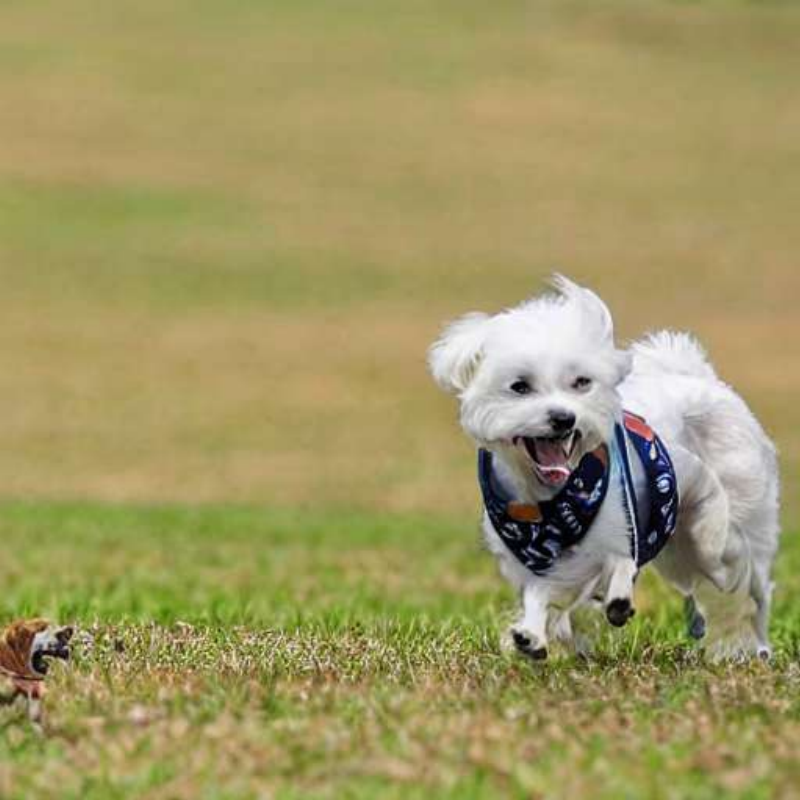}
  }
  \subfloat
  {
      \includegraphics[width=0.18\textwidth]{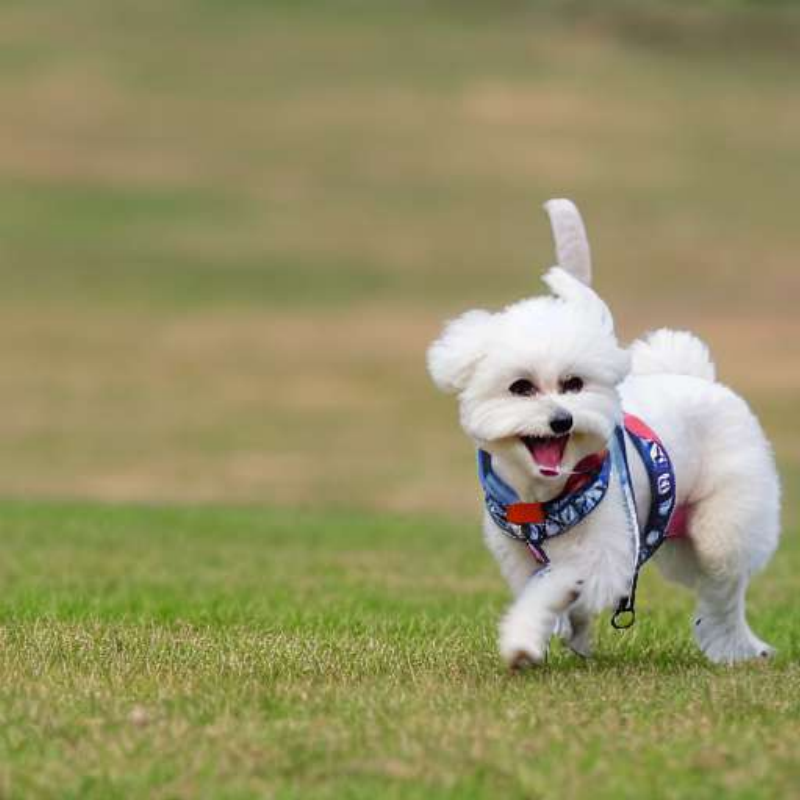}
  }
  \subfloat
  {
      \includegraphics[width=0.18\textwidth]{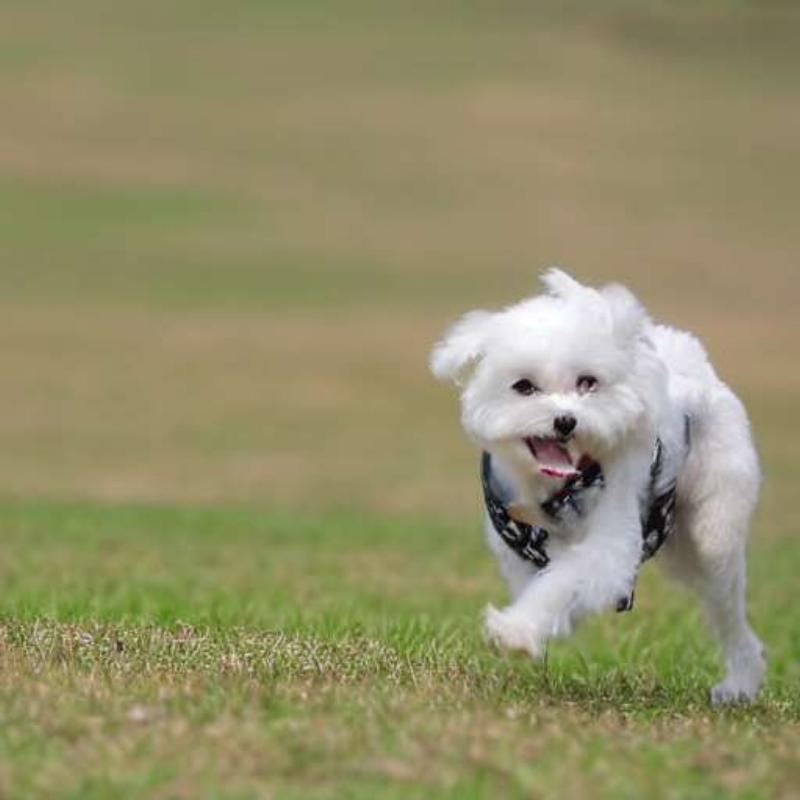}
  }
  \subfloat
  {
      \includegraphics[width=0.18\textwidth]{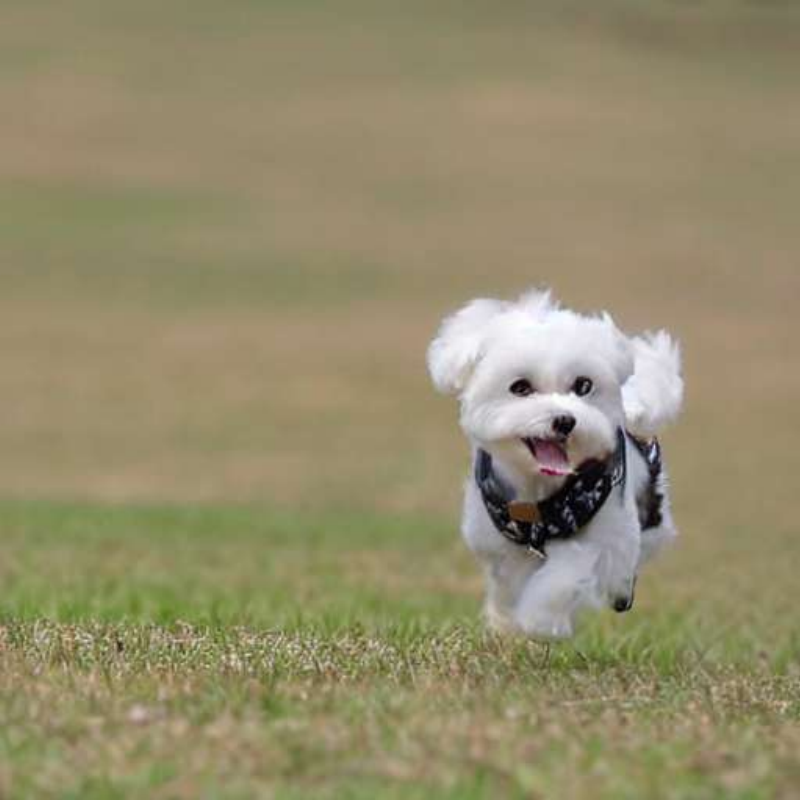}
  }
  
  \rotatebox{90}{\normalsize{~~~~~~~~~\textit{A sitting cat}}}
  \hspace{-3pt}
  \subfloat[(a) Input]
  {
      \includegraphics[width=0.18\textwidth]{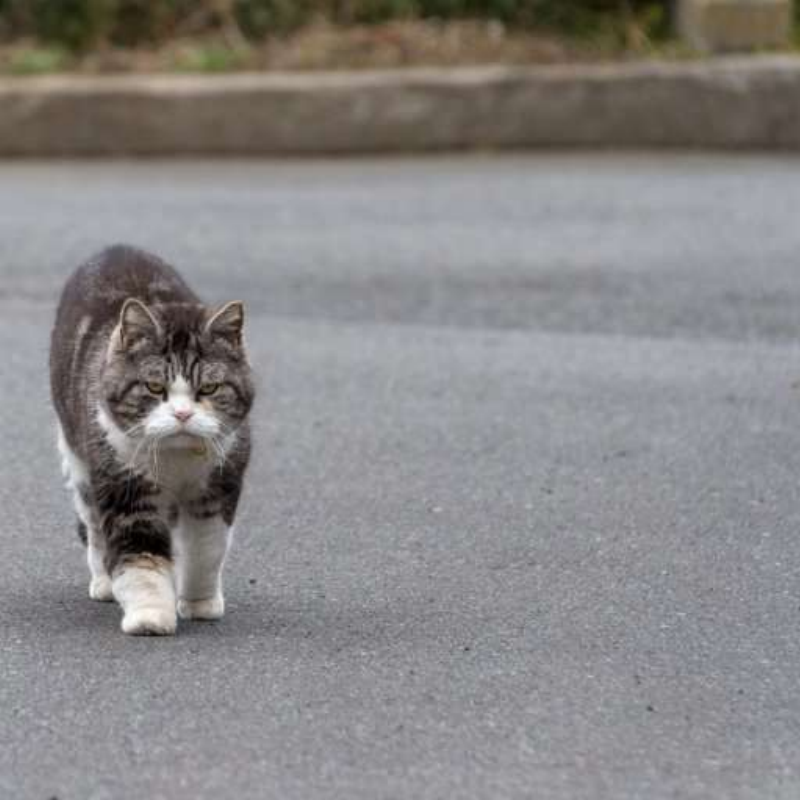}
  }
  \subfloat[(b) InstructPix2Pix]
  {
      \includegraphics[width=0.18\textwidth]{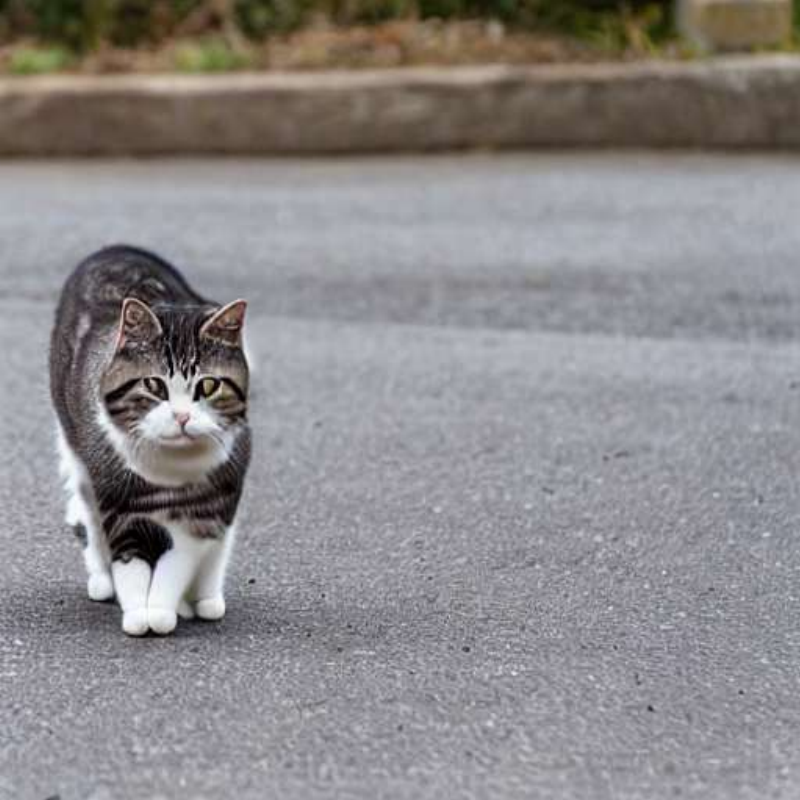}
  }
  \subfloat[~(c) PnP]
  {
      \includegraphics[width=0.18\textwidth]{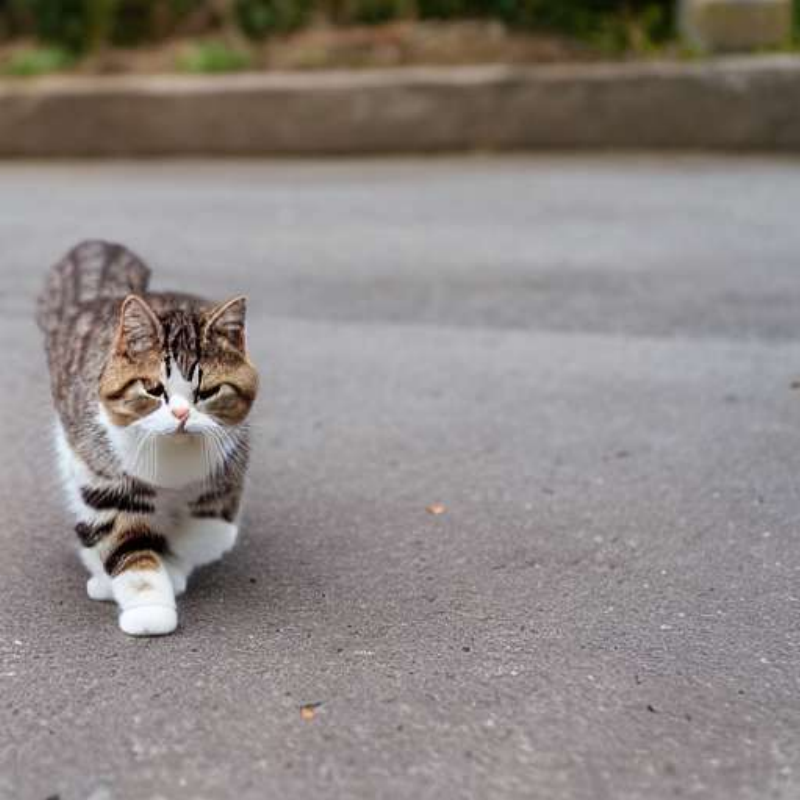}
  }
  \subfloat[~(d) MasaCtrl]
  {
      \includegraphics[width=0.18\textwidth]{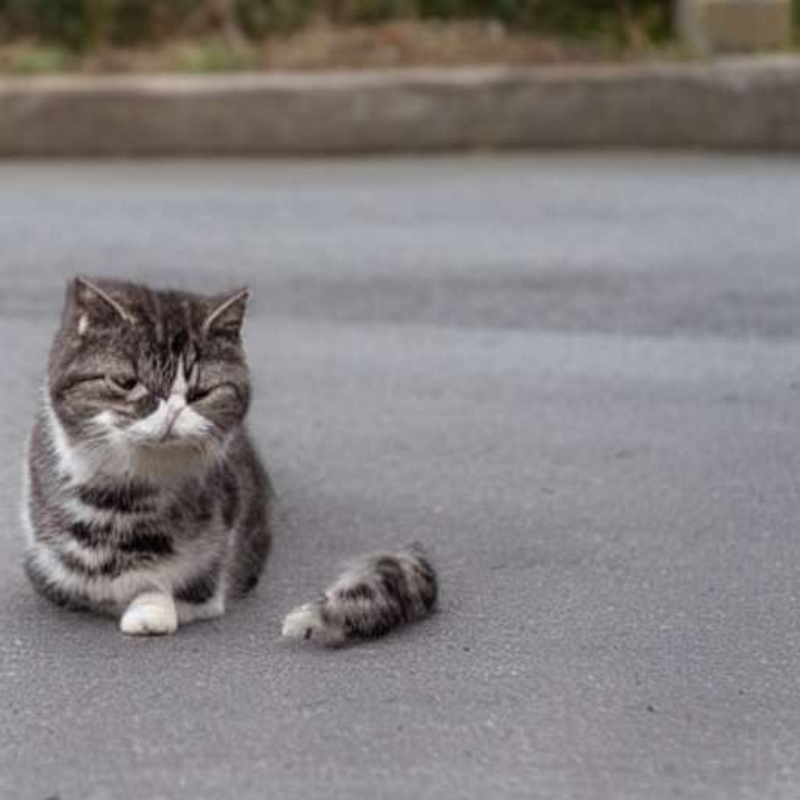}
  }
  \subfloat[~~(e) Move\&Act\newline\centering(Ours)]
  {
      \includegraphics[width=0.18\textwidth]{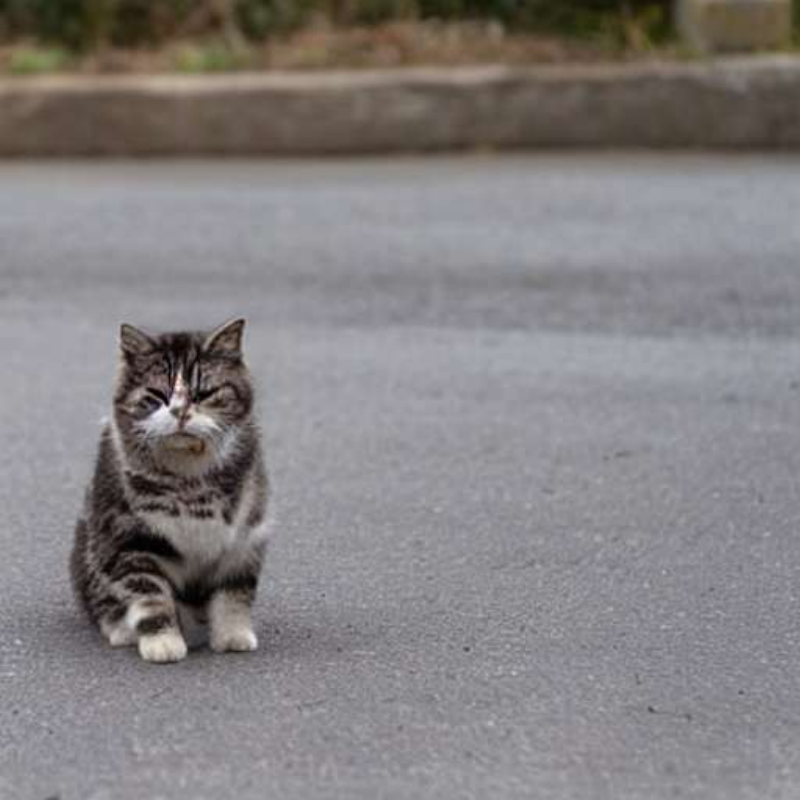}
  }
  \vspace{-0.5em}
  \caption{Visual comparison of different methods for modifying only actions.}
  \vspace{-1.0em}
  \label{fig7}
\end{figure*}

\subsection{Implementation Details}
In the experimental section, we employ the stable diffusion~\cite{ref5} using checkpoint v1.4. We utilize a 50-step DDIM inversion process, and the classifier-free guidance is set at 7.5. During the 35-$th$ step of inversion, we update the latent code. The hyperparameter values for $ \lambda_{oii}$, $ \lambda_{sai}$, and $ \lambda_{bg}$ in Eq.~\eqref{eq:inversion-loss} are set to 0.5, 0.25, and 0.25, respectively. In order to update the latent code in accordance with Eq.~\eqref{eq:update}, we establish the step size $\alpha$ at 150 and designate the number of iterations as 50. From the 7-$th$ step of the reverse process onward, we substitute the key and value within the self attention mechanism as per Eq~\eqref{eq:kv-exchange}. All experiments are conducted on a single 3090 GPU, well demonstrating the democratization of our introduced method in this paper.

\subsection{Evaluation Dataset}
We manually annotate 200 condition-image pairs from the web, the collected data focuses on non-rigid objects editing. Each image is linked to a specific condition, which includes the inversion prompt, editing prompt, and bounding box. Inversion prompts follow a common template: ``A photo of $<$object$>$'', \emph{e.g.}, ``A photo of cat''. The editing prompt template is ``A $<$action$>$ $<$object$>$'', \emph{e.g.}, ``A running cat''. The bounding box designates the object's generation location. 

\subsection{Qualitative Result}

The image editing results of our method are vividly depicted in Figure\,\ref{fig6}. Our approach exhibits notable strengths in three key areas: (1) the edited object accurately adheres to the instructed action; (2) the edited object precisely aligns with the designated position box; (3) the background is seamlessly integrated into the edited image. As such, our method significantly enhances the manipulation of objects and the preservation of background integrity during the image editing process. This ultimately results in a more refined and visually appealing output, demonstrating the efficacy of our proposed technique.

We configure the generated position of the edited object to match the object's original location, for comparing with existing methods. As shown in Figure\,\ref{fig1} and Figure\,\ref{fig7}, our method exhibits superior background preservation, and the actions of the edited objects align more closely with the user-provided prompts. This comparison highlights our method in achieving enhanced consistency.

\subsection{Quantitative Result}

We calculate the following metrics to compare with previous methods: (1) CLIP Score~\cite{ref25}, which assesses the association between text and images. 
A higher CLIP Score value signifies a greater correlation between text-image pairs. (2) YOLO Score~\cite{ref26}, which evaluates the accuracy of the generated position of the edited object using the $AP_{50}$ metric. To detect the bounding box and predict the category of the edited object in the generated image, we employ YOLOv8~\cite{ref27}.

As illustrated in Table~\ref{table1}, our Move\&Act achieves a higher CLIP Score, indicating that the editing results produced by our method exhibit greater consistency with the user-provided prompts. We also report our $AP_{50}$ score. Since the methods we compared do not possess the capability to control the generated position of the edited object, the $AP_{50}$ values for these methods are left blank.

\begin{table}[!t]
  \centering
  \setlength{\tabcolsep}{10pt}
  \renewcommand\arraystretch{1.2}
  \begin{tabular}{@{}lcc@{}}
    \toprule
    Method & CLIP-Score & $AP_{50}$ \\
    \midrule
    InstructPix2Pix & 0.2966 $\pm$ 0.0013 & - \\
    PnP & 0.3035 $\pm$ 0.0009 & - \\
    MasaCtrl & 0.3066 $\pm$ 0.0008 & - \\
    Move\&Act (Ours) & \textbf{0.3098 $\pm$ 0.0011} & \textbf{33.70 $\pm$ 0.74} \\
  \bottomrule
  \end{tabular}
  \caption{A comparison of quantitative results.}
  \label{table1}
\end{table}

\begin{figure}[!t]
  \centering
  \captionsetup[subfloat]{labelsep=none,format=plain,labelformat=empty}
  \rotatebox{90}{\scriptsize{~~Attention maps}}
  \hspace{-7pt}
  \subfloat   
  {
      \includegraphics[width=0.22\linewidth]{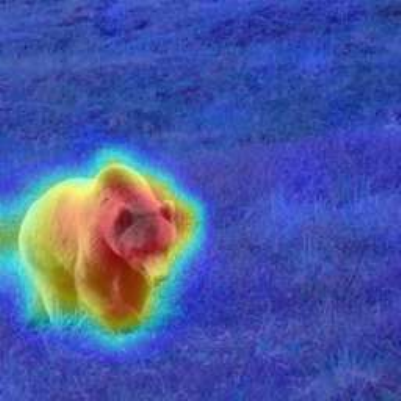}
  }
  \subfloat
  {
      \includegraphics[width=0.22\linewidth]{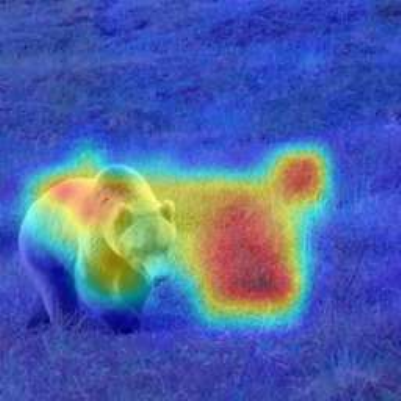}
  }
  \subfloat
  {
      \includegraphics[width=0.22\linewidth]{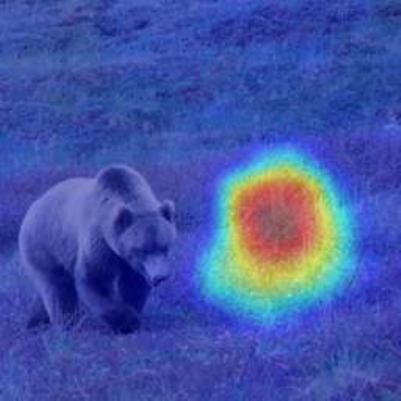}
  }
  \subfloat
  {
      \includegraphics[width=0.22\linewidth]{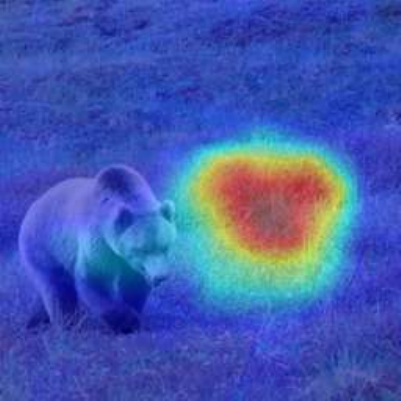}
  }

  \rotatebox{90}{\scriptsize{~~~Editing results}}
  \hspace{-7pt}
  \subfloat[(a) Input]   
  {
      \includegraphics[width=0.22\linewidth]{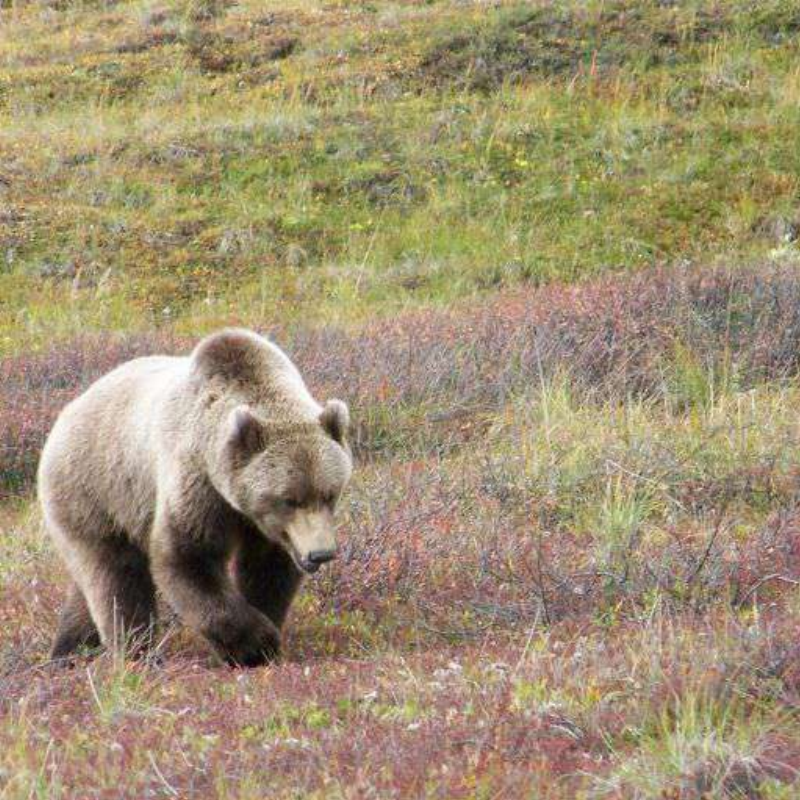}
  }
  \subfloat[(b) Step 25]
  {
      \includegraphics[width=0.22\linewidth]{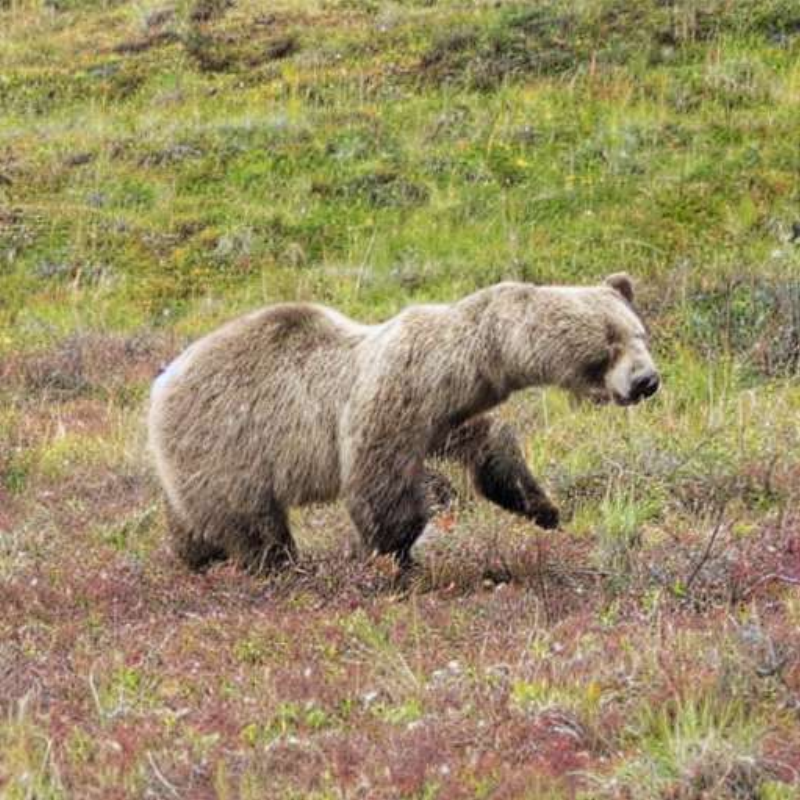}
  }
  \subfloat[(c) Step 35]
  {
      \includegraphics[width=0.22\linewidth]{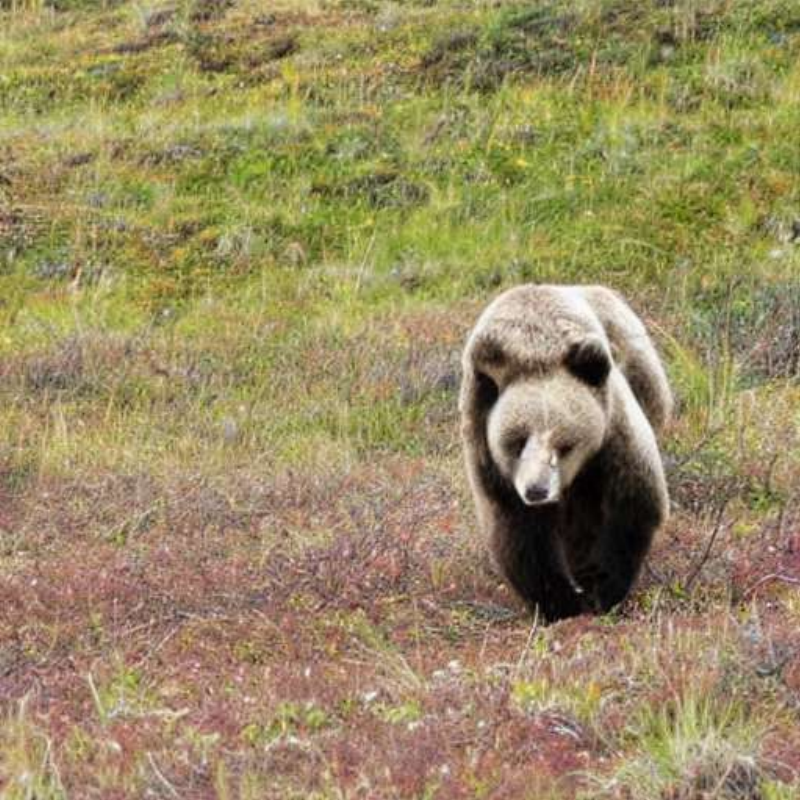}
  }
  \subfloat[(d) Step 45]
  {
      \includegraphics[width=0.22\linewidth]{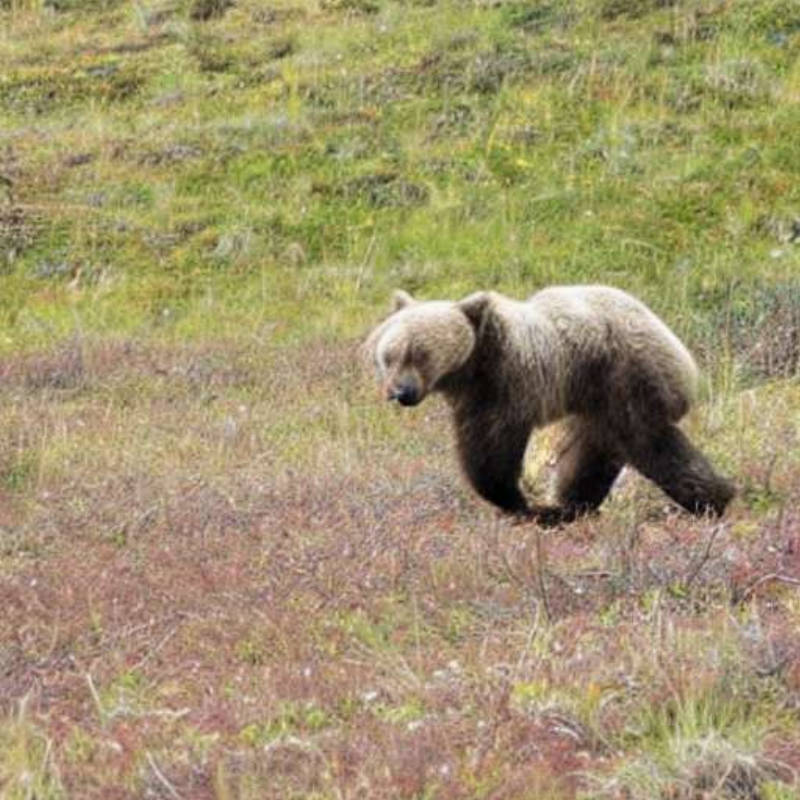}
  }
  \caption{Editing results of different time step settings to update $z_{t}$ during inversion.}
  \label{fig8}
\end{figure}

\subsection{Ablation Study}

\subsubsection{Time Step to Update $z_{t}$ During Inversion.}
To determine the optimal time step for updating the latent code during inversion branch in Eq.~\eqref{eq:inversion-loss}, we experiment with the 25-$th$, 35-$th$, and 45-$th$ time steps. As illustrated in Figure\,\ref{fig8}, setting the time step to 25 results in incomplete object information transfer, causing the object not to appear in the target area. Conversely, a time step of 45 allows for object information transfer but yields unsatisfactory visual results. The attention map visualization reveals that updating the latent code at the 35-$th$ time step achieves the most effective object information transfer.

\begin{figure}[!t]
  \centering
  \subfloat[Input]   
  {
      \includegraphics[width=0.23\linewidth]{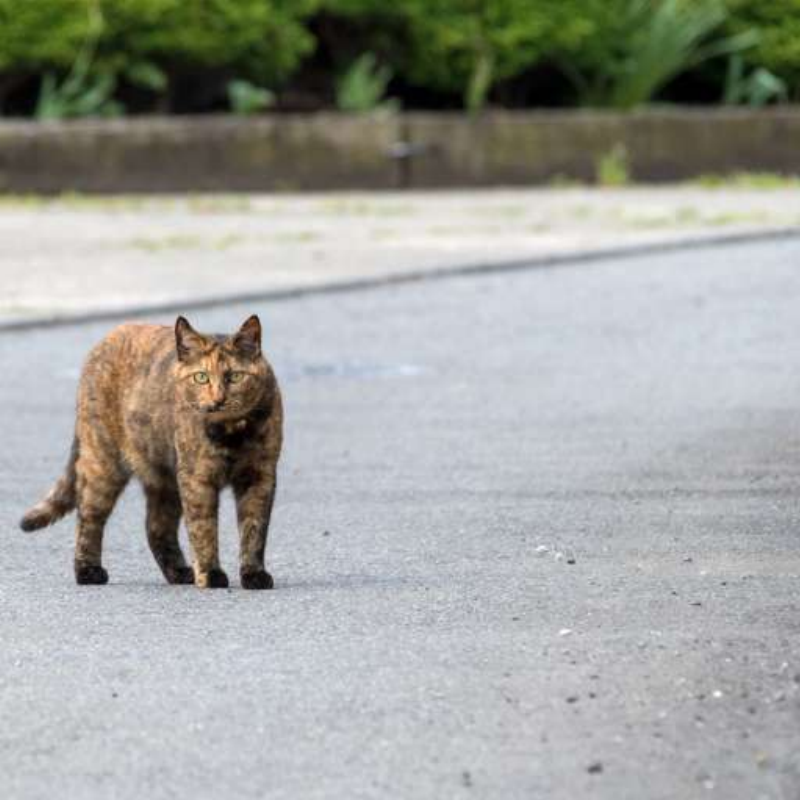}
  }
  \subfloat[25 iterations]
  {
      \includegraphics[width=0.23\linewidth]{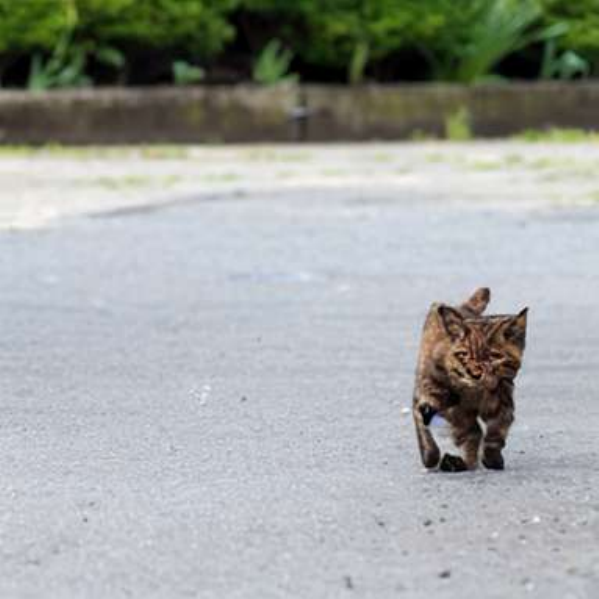}
  }
  \subfloat[50 iterations]
  {
      \includegraphics[width=0.23\linewidth]{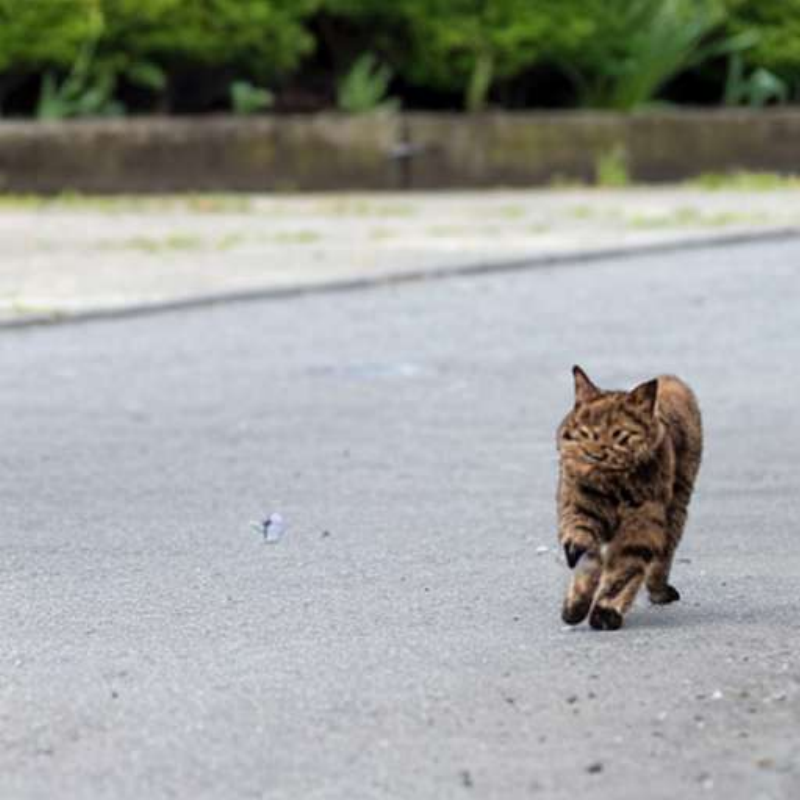}
  }
  \subfloat[75 iterations]
  {
      \includegraphics[width=0.23\linewidth]{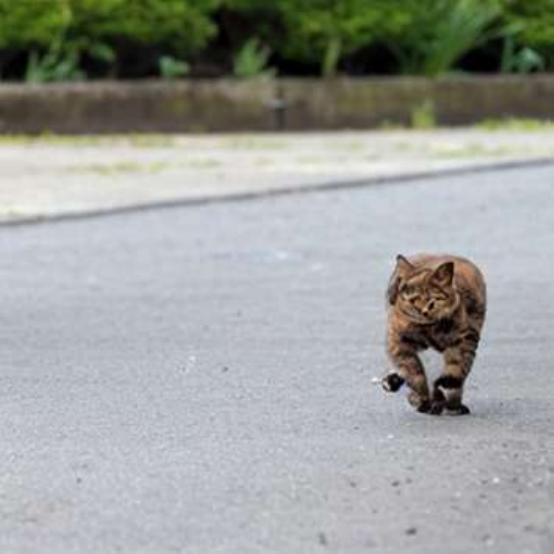}
  }
  \caption{Editing results of different iteration settings.}
  \label{fig9}
\end{figure}

\subsubsection{Iterations of Latent Updating.}
We perform experiments to find the suitable number of iterations to converge Eq.~\eqref{eq:inversion-loss}, as shown in Figure\,\ref{fig9}. With 25 iterations, the edited image struggles to maintain the original background effectively, and the cat's action appears distorted. Conversely, setting the iteration number to a higher value of 75 allows for better preservation of the original background; however, the cat's action does not align well with the user's provided prompt.

Taking these observations into account, we opt for an iteration number of 50 steps. This setting strikes a balance between preserving the background and accurately editing the object's action, resulting in a more visually appealing and coherent edited image that adheres to the editing prompt.

\subsubsection{Starting Time Step for Editing Consistency.}
\begin{figure}[!t]
  \centering
  \subfloat[Input]   
  {
      \includegraphics[width=0.3\linewidth]{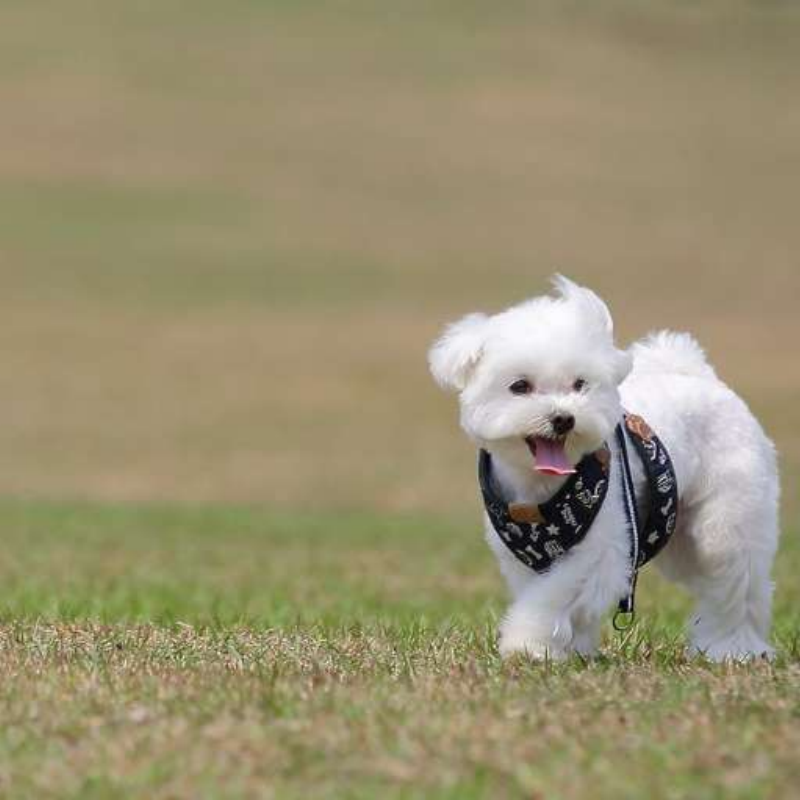}
  }
  \subfloat[Step 0]
  {
      \includegraphics[width=0.3\linewidth]{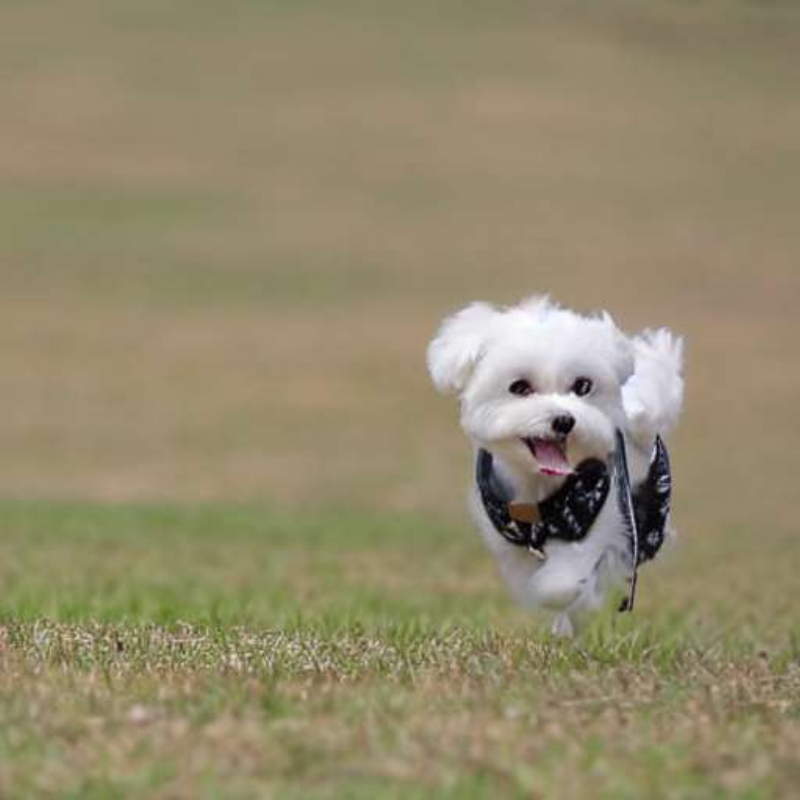}
  }
  \subfloat[Step 7]
  {
      \includegraphics[width=0.3\linewidth]{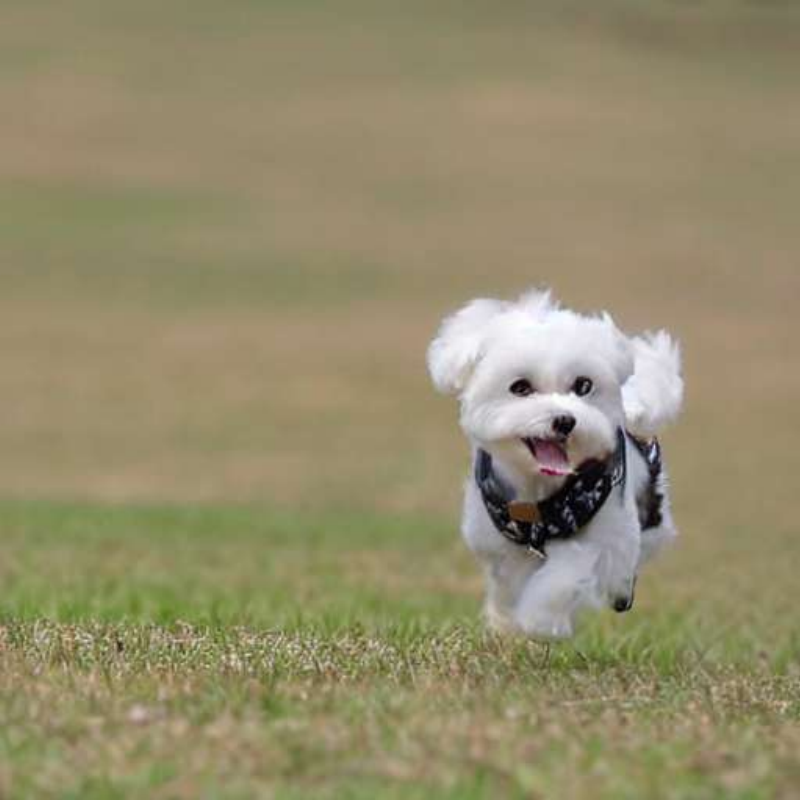}
  }
  
  \subfloat[Step 15]
  {
      \includegraphics[width=0.3\linewidth]{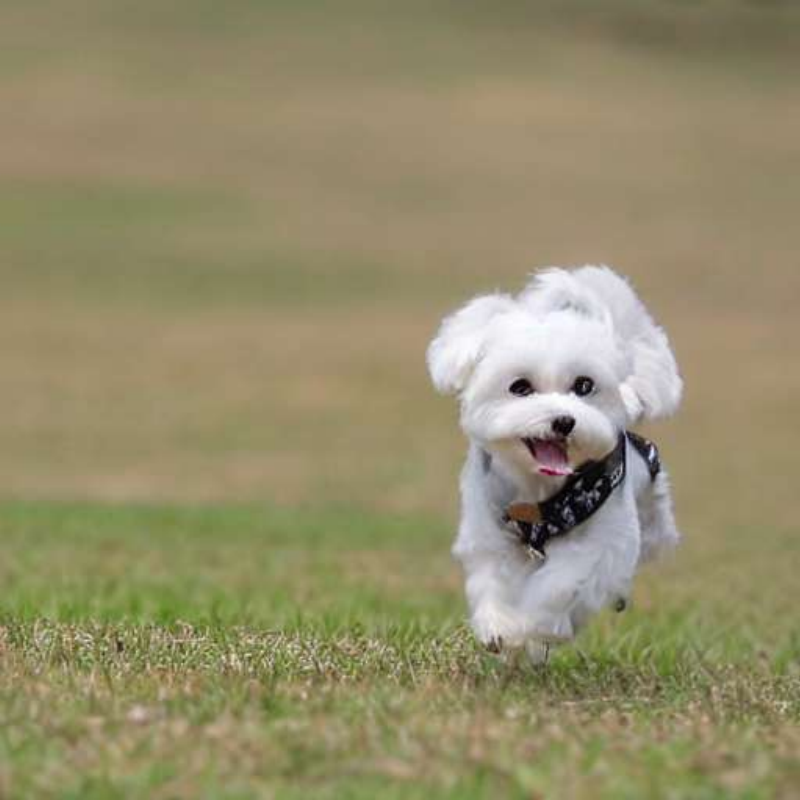}
  }
  \subfloat[Step 30]
  {
      \includegraphics[width=0.3\linewidth]{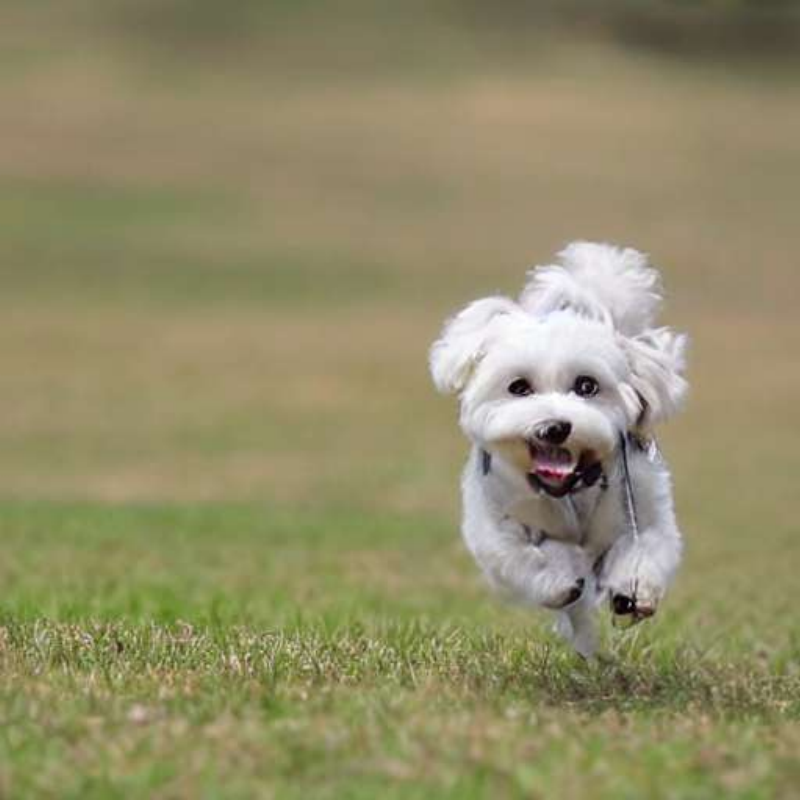}
  }
  \subfloat[Step 45]
  {
      \includegraphics[width=0.3\linewidth]{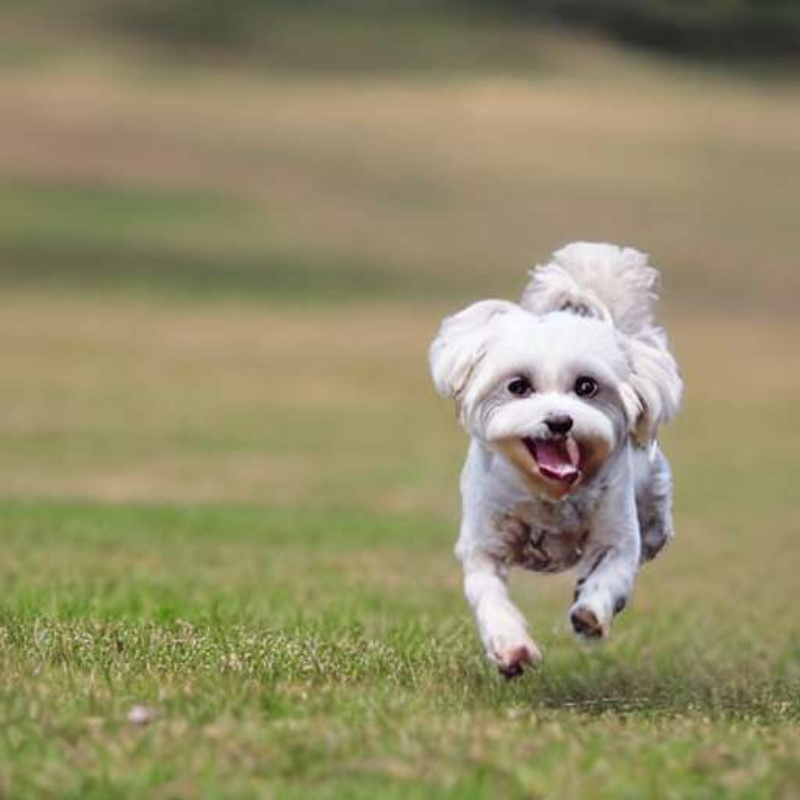}
  }
  \caption{Time step for editing consistency.}
  \label{fig10}
\end{figure}

In Eq.~\eqref{eq:kv-exchange}, our approach utilizes image features in self-attention to query the key and value at the corresponding time step during inversion, starting from time step $S$ in the reverse process. As depicted in Figure\,\ref{fig10}, we experimented with different values of $S$. Setting $S$ to a larger time step results in the edited object losing its original appearance, with alterations in the dog's decorations, hair, and the image background. When $S$ is set to 7, the editing outcome aligns more closely with the user's prompt.

\subsubsection{Self-attention Layer Setting.}
\begin{figure}[!t]
  \centering
  \subfloat[Input]   
  {
      \includegraphics[width=0.23\linewidth]{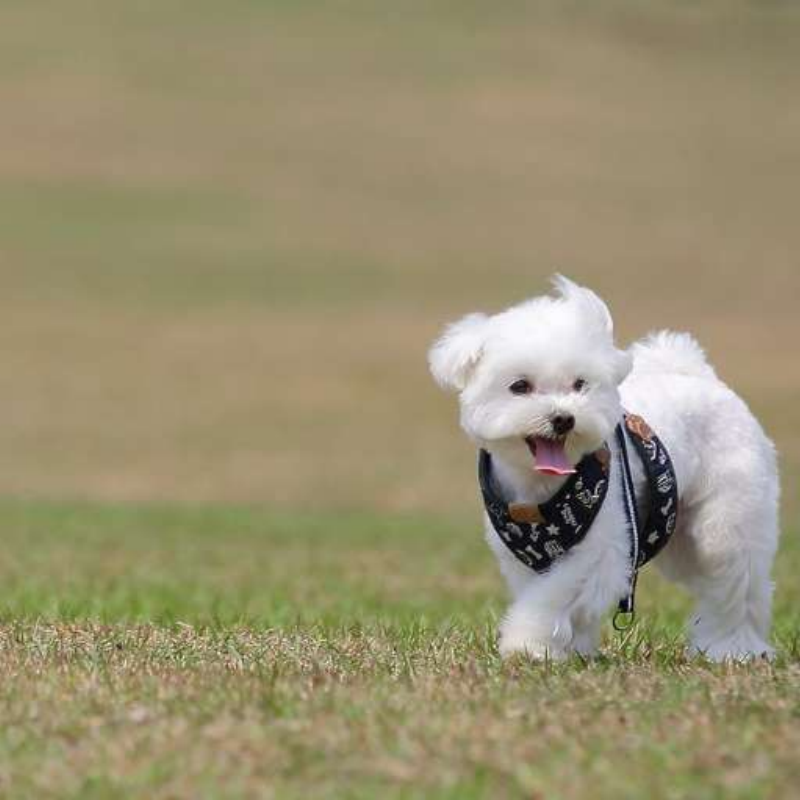}
  }
  \subfloat[Whole UNet]
  {
      \includegraphics[width=0.23\linewidth]{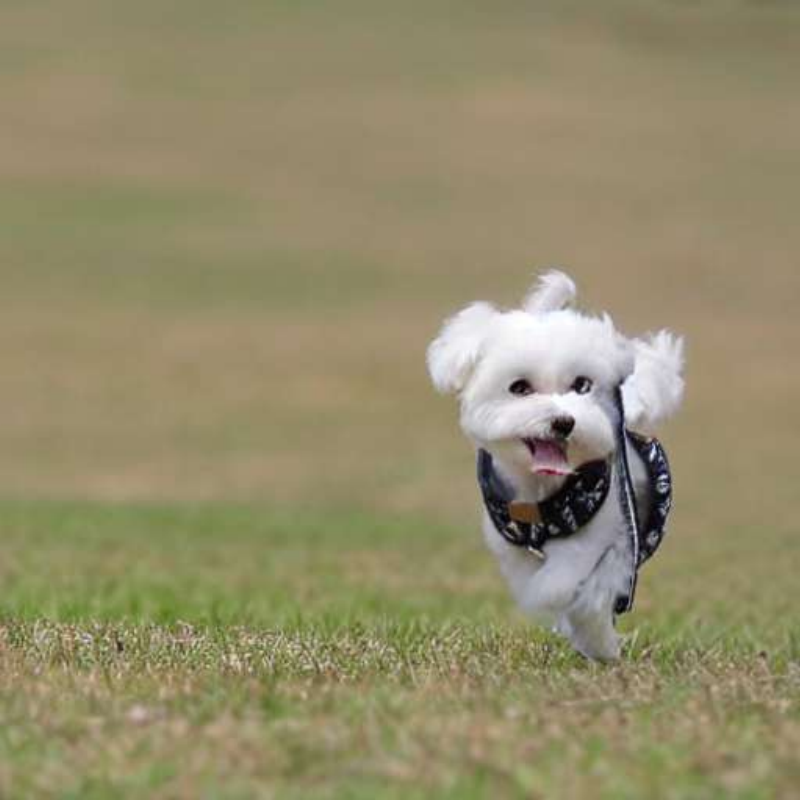}
  }
  \subfloat[Encoder]
  {
      \includegraphics[width=0.23\linewidth]{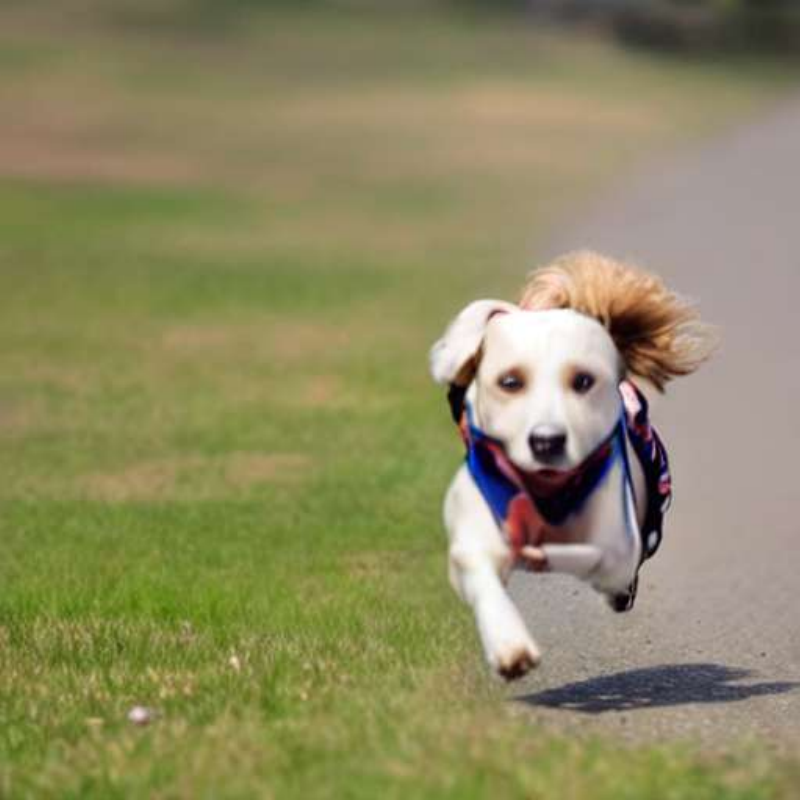}
  }
  \subfloat[Decoder]
  {
      \includegraphics[width=0.23\linewidth]{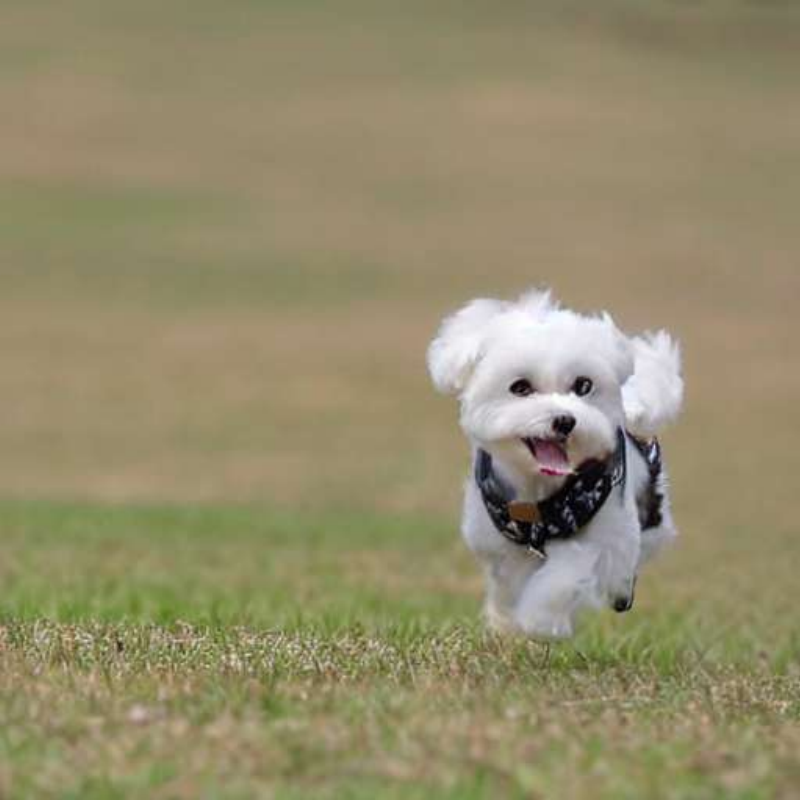}
  }
  \caption{Results of different self-attention layer settings.}
  \label{fig11}
\end{figure}

We only replace the key and value in the specified self attention layer in Eq.~\eqref{eq:kv-exchange}. In order to test which self-attention layers in UNet play a major role in restoring the object appearance and the image background, we implement an experiment and test three replacement schemes, which were to replace all self-attention key and value in UNet, replace only self-attention key and value of the UNet encoder and replace only self-attention key and value of the UNet decoder. As shown in Figure\,\ref{fig11}, decoder self-attention plays a major role in restoring the object appearance and the image background, while encoder self-attention plays a minor role. Thus, we only replace the key and value of the UNet decoder's self attention.

\section{Limitation and Future Work}
Complex image input is a challenge for our method, particularly in effectively restoring the source area’s background when it is surrounded by complex background. Additionally, the kernel size choice for the dilation operator is pivotal in our approach. A smaller kernel size results in an edge area with deficient features, leading to suboptimal outcomes. Conversely, an excessively large kernel size produces a broad edge area, encompassing background regions too distant from the source area to be relevant for restoration. To refine our method, future research will concentrate on investigating source area inpainting loss with an enhanced capacity for background recovery, thereby ensuring more precise and consistent results across various editing scenarios.

\section{Conclusion}
We have introduced a tuning-free approach for achieving consistent image editing, enabling users to  modify an object's action and control its generated position, while ensuring superior background preservation. Our method transfers object information from the source area to the target area during the inversion stage and employs the features of the UNet denoiser to restore the source area's background, preserving the unedited area's background. This innovative approach facilitates more precise control over image objects and enhances background consistency before and after editing, causing a seamless and visually appealing outcome.

\section*{Acknowledgement}
This work was supported by National Science and Technology Major Project (No. 2022ZD0118201), the National Science Fund for Distinguished Young Scholars (No.62025603), the National Natural Science Foundation of China (No. U21B2037, No. U22B2051, No. U23A20383, No. U21A20472, No. 62176222, No. 62176223, No. 62176226, No. 62072386, No. 62072387, No. 62072389, No. 62002305 and No. 62272401), and the Natural Science Foundation of Fujian Province of China (No. 2021J06003, No.2022J06001).


\end{document}